\documentclass[sigconf]{acmart}

\author{Souvika Sarkar}
\email{szs0239@auburn.edu}
\authornote{Both authors contributed equally to this work.}
\affiliation{%
  \institution{Auburn University}
  \streetaddress{}
  \city{}
  \state{Alabama}
  \country{USA}
  \postcode{}
}

\author{Mohammad Fakhruddin Babar}
\email{m.babar@wsu.edu}
\authornotemark[1]
\affiliation{%
  \institution{Washington State University}
  \streetaddress{}
  \city{Pullman}
  \country{USA}}

\author{Md Mahadi Hassan}
\email{mzh0167@auburn.edu}
\affiliation{%
  \institution{Auburn University}
  \streetaddress{}
  \city{}
  \state{Alabama}
  \country{USA}
  \postcode{}
}

\author{Monowar Hasan}
\email{monowar.hasan@wsu.edu}
\affiliation{%
  \institution{Washington State University}
  \streetaddress{}
  \city{Pullman}
  \country{USA}}

\author{Shubhra Kanti Karmaker Santu}
\email{sks0086@auburn.edu}
\affiliation{%
  \institution{Auburn University}
  \streetaddress{}
  \city{}
  \state{Alabama}
  \country{USA}
  \postcode{}
}

\usepackage[mathletters]{ucs}

\usepackage{adjustbox}
\usepackage{booktabs}
\usepackage{longtable}
\usepackage{makecell}
\usepackage{caption}
\usepackage{amsmath}
\usepackage{xspace}
\usepackage{xcolor}
\usepackage{comment}
\usepackage{enumitem}
\newtheorem{definition}{Definition}

\usepackage{multirow}
\usepackage{rotating}
\usepackage{soul}
\usepackage{lipsum}
\usepackage{pifont}  
\usepackage{dcolumn}

\usepackage{colortbl}

\usepackage{marvosym}
\usepackage{fontawesome}
\usepackage{csquotes}

\usepackage{latexsym}
\usepackage{ulem}
\usepackage[most]{tcolorbox} 

\newcommand{\eg}{{e.g.,}\xspace}
\newcommand{\viz}{{viz.,}\xspace}

\newcommand{\ie}{{i.e.,}\xspace}

\newcommand{\ci}{{\it (i) }}
\newcommand{\cii}{{\it (ii) }}
\newcommand{\ciii}{{\it (iii) }}

\newcommand{\ca}{{\it (a) }}
\newcommand{\cb}{{\it (b) }}
\newcommand{\cc}{{\it (c) }}
\newcommand{\cd}{{\it (d) }}
\newcommand{\ce}{{\it (e) }}
\newcommand{\cf}{{\it (f) }}

\definecolor{darkspringgreen}{rgb}{0.09, 0.45, 0.27}

\definecolor{notecolor}{rgb}{0.8,0,0} 

\newcommand{\acmbfpar}[1]{{\vspace*{0.3\baselineskip}\noindent\bfseries #1\quad}}

\AtBeginDocument{%
  \providecommand\BibTeX{{%
    \normalfont B\kern-0.5em{\scshape i\kern-0.25em b}\kern-0.8em\TeX}}}

\newcommand{\clrgray}{\cellcolor{gray!50}}
\newcommand{\hardware}{Raspberry Pi, Jetson, UP\textsuperscript{2}, and UDOO\xspace}

\newcommand{\rpi}{Raspberry Pi\xspace}
\newcommand{\rpib}{Raspberry Pi 4 Model B\xspace}
\newcommand{\jet}{Jetson\xspace}
\newcommand{\jetn}{Jetson Nano\xspace}
\newcommand{\up}{UP\textsuperscript{2}\xspace}
\newcommand{\udo}{UDOO\xspace}
\newcommand{\udob}{UDOO Bolt\xspace}

\copyrightyear{2024} 
\acmYear{2024} 
\setcopyright{acmlicensed}\acmConference[ICPE '24]{Proceedings of the 15th ACM/SPEC International Conference on Performance Engineering}{May 7--11, 2024}{London, United Kingdom}
\acmBooktitle{Proceedings of the 15th ACM/SPEC International Conference on Performance Engineering (ICPE '24), May 7--11, 2024, London, United Kingdom}
\acmDOI{10.1145/3629526.3645054}
\acmISBN{979-8-4007-0444-4/24/05}

\renewcommand{\shortauthors}{S. Sarkar, M. Babar, M. Hassan, M. Hasan, S. Santu}

\begin{document}

\title[Processing Natural Language on Embedded Devices]{Processing Natural Language on Embedded Devices: How Well Do Transformer Models Perform?}

\begin{abstract}

Voice-controlled systems are becoming ubiquitous in many IoT-specific applications such as home/industrial automation, automotive infotainment, and healthcare. While cloud-based voice services (\eg Alexa, Siri) can leverage high-performance computing servers, some use cases (\eg robotics, automotive infotainment) may require to execute the natural language processing (NLP) tasks offline, often on resource-constrained embedded devices. Transformer-based language models such as BERT and its variants are primarily developed with compute-heavy servers in mind. Despite the great performance of BERT models across various NLP tasks, their large size and numerous parameters pose substantial obstacles to offline computation on embedded systems. Lighter replacement of such language models (\eg DistilBERT and TinyBERT) often sacrifice accuracy, particularly for complex NLP tasks. Until now, it is still unclear \ca whether the state-of-the-art language models, \viz BERT and its variants are deployable on embedded systems with a limited processor, memory, and battery power and \cb if they do, what are the ``right'' set of configurations and parameters to choose for a given NLP task. 

This paper presents a \textit{performance study of transformer language models} under different hardware configurations and accuracy requirements and derives empirical observations about these resource/accuracy trade-offs. In particular, we study how the most commonly used BERT-based language models (\viz BERT, RoBERTa, DistilBERT, and TinyBERT) perform on embedded systems. We tested them on \textbf{\textit{four}} off-the-shelf embedded platforms (\hardware) with 2 GB and 4 GB memory (\ie a total of \textbf{\textit{eight}} hardware configurations) and \textbf{\textit{four}} datasets (\ie HuRIC, GoEmotion, CoNLL, WNUT17) running various NLP tasks. Our study finds that executing complex NLP tasks (such as ``sentiment'' classification) on embedded systems is \textit{feasible} even without any GPUs (\eg \rpi with 2 GB of RAM). 
Our findings can help designers understand the deployability and performance of transformer language models, especially those based on BERT architectures.

\end{abstract}



\settopmatter{printfolios=true}
\maketitle

\section{Introduction}\label{introduction}


The natural language processing (NLP) domain and the emergence of large language models rapidly transform how we interact with technology. With the proliferation of IoT-specific applications, the demand for voice-controlled services that can perform tasks by responding to spoken commands is growing. Use cases of NLP applications, especially those that need embedded and mobile systems, include home automation, healthcare, industrial control, and automotive infotainment. To design a dialogue-based interaction system for a target device, we need models that are feasible to~\ca run on the hardware and~\cb meet the desired level of accuracy. Although some services such as digital assistants (\eg Alexa, Siri, Cortana) may leverage cloud resources for processing human voices, there exist applications (\eg offline home/industrial robots, automotive infotainment, battlefield/military equipment) that may not have network connectivity, thus require to synthesize NLP tasks on the embedded device itself. The challenge is understanding the \textit{feasibility} of running a large language model on \textit{resource-limited} devices.

Transformer-based architectures~\cite{vaswani2017attention}, especially BERT-based models~\cite{bert}, have established themselves as popular state-of-the-art baselines for many NLP tasks, including \textit{Intent Classification (IC)}, \textit{Sentiment Classification (SC)}, and \textit{Named Entity Recognition (NER)}. However, a well-known criticism of BERT-based architectures is that they are data-hungry and consume a lot of memory and energy; therefore, deploying them in embedded systems is challenging. In fact, due to their excessive size (431~MB) and parameters (110~M), deploying a pre-trained BERT model (called $BERT_{Base}$) in resource-constrained embedded devices is often impractical, especially at the production level with certain minimum accuracy/performance requirements. Lighter versions of BERT (\eg~DistilBERT~\cite{distilbert} and TinyBERT~\cite{tinybert}) often result in accuracy losses. The degree of degradation in performance depends on the difficulty of the task, especially since those models often cannot perform well on complex NLP tasks, including emerging entity \cite{wnut17} or mixed emotion detection~\cite{goemotions}. Therefore, designers must make an inevitable trade-off between an accurate model and one that can run smoothly in a resource-constrained environment. Unfortunately, developers often have little idea about this trade-off and have to spend a lot of time conducting trial-and-error experiments to find a suitable architecture that is feasible for the target (resource-constrained) hardware and meets a desired level of accuracy.

From a developer's perspective, it is still unclear what is the ``right'' BERT-based architecture to use for a given NLP task that can strike a suitable \textit{trade-off between the resources available and the minimum accuracy} desired by the user. Due to the staggering size of the $BERT_{Base}$ model, we experiment with different ``distilled'' BERT models (\eg DistilBERT and TinyBERT) for IC, SC, and NER tasks. However, existing ready-to-use distilled models perform poorly on some SC and NER datasets (Sec.~\ref{results}). Hence, there is a need to \textit{explore other models that can better optimize the efficiency/accuracy trade-offs}.

This research performs an \textit{exploratory study of BERT-based models\footnote{\ul{Note:} there exist other large language models, such as GPT~\cite{DBLP:journals/corr/abs-2005-14165} from OpenAI and LaMDA~\cite{DBLP:journals/corr/abs-2201-08239} from Google. However, they are even more resource-hungry than BERT (thus less suitable for embedded deployment), and some are close-sourced. Hence, our initial study limits on BERT-based architectures.} under different resource constraints and accuracy budgets to derive empirical data about these resource/accuracy trade-offs}. We aim to answer the following questions: \ca how can we determine the suitable BERT architecture that runs on a target hardware and meets user-defined performance requirements (accuracy, inference time)? \cb what are the trade-offs between accuracy and model size as we perform optimizations (such as \textit{pruning}) to run them on limited embedded device memory? and \cc what are the implications of performing pruning on accuracy and corresponding resource usage, including memory, inference time, and energy consumption? In answer to those questions, we observe the overhead of running various BERT architectures on four different hardware (\viz~\rpi~\cite{raspberrypi}, \jetn~\cite{jetsonnano}, \up~\cite{upboard}, and \udo~\cite{udoobolt}). Our experiments suggest that some BERT models (specifically those that are ``distilled'') failed to achieve desired performance goals (\eg F1 score) for various NLP tasks. Further, although pruning can reduce model size, it does not significantly help in energy efficiency.

\acmbfpar{Contributions.} Our study fills the gap between simulation-based studies and real-world scenarios, as no prior work has deployed these models on embedded platforms.  The findings of this work can help designers choose alternative BERT-based architectures under given resource constraints, thus saving development time and hassle. To ensure reproducibility, our implementation and related documentation is~\textbf{\ul{publicly available}}~\cite{repo_link}. 

We made the following contributions in this paper.


\begin{itemize}[itemsep=0.4ex,partopsep=0ex,parsep=0ex]


    \item Our study systematically investigates the performance of BERT-based language models on four off-the-shelf embedded platforms (\hardware) with two different memory variants (2 GB and 4 GB RAMs). We analyzed the trade-offs between complexity and accuracy across multiple NLP tasks. (Sec.~\ref{exp_design}-Sec.~\ref{results}).

    \item We explore the feasibility of deploying complex NLP tasks on embedded systems and analyze them under three \textit{metrics}: \ca inference time, \cb memory usage, and \cc energy consumption. We developed a lookup table through empirical observations that will be useful for system designers to decide suitable model configurations for the target platform (Sec.~\ref{results}).


\end{itemize}

\acmbfpar{Our key findings.} The observations from executing the NLP tasks on our test platforms are as follows: \ca simpler NLP tasks such as IC can result in a relatively high (90\%+) F1 score; \cb the time required to perform inference proportional to the size of the trained model and trimming them result in some accuracy loss (\eg 60\% of reduction in model size could reduce the accuracy by  50\%); \cc the energy consumption on our test hardware remains relatively consistent, whether we prune the model or not; and \cd while GPUs play a role in decreasing inference time, the unavailability of GPUs can be compensated by faster CPUs.


 

We now start with 
the problem statement and present selected datasets (Sec.~\ref{problem_statement}). Section~\ref{exp_design} describes our experiment setup before we discuss our findings in Sec.~\ref{results}.

\begin{figure*}[!t]
\centering
        \includegraphics[width=0.6850\linewidth,]{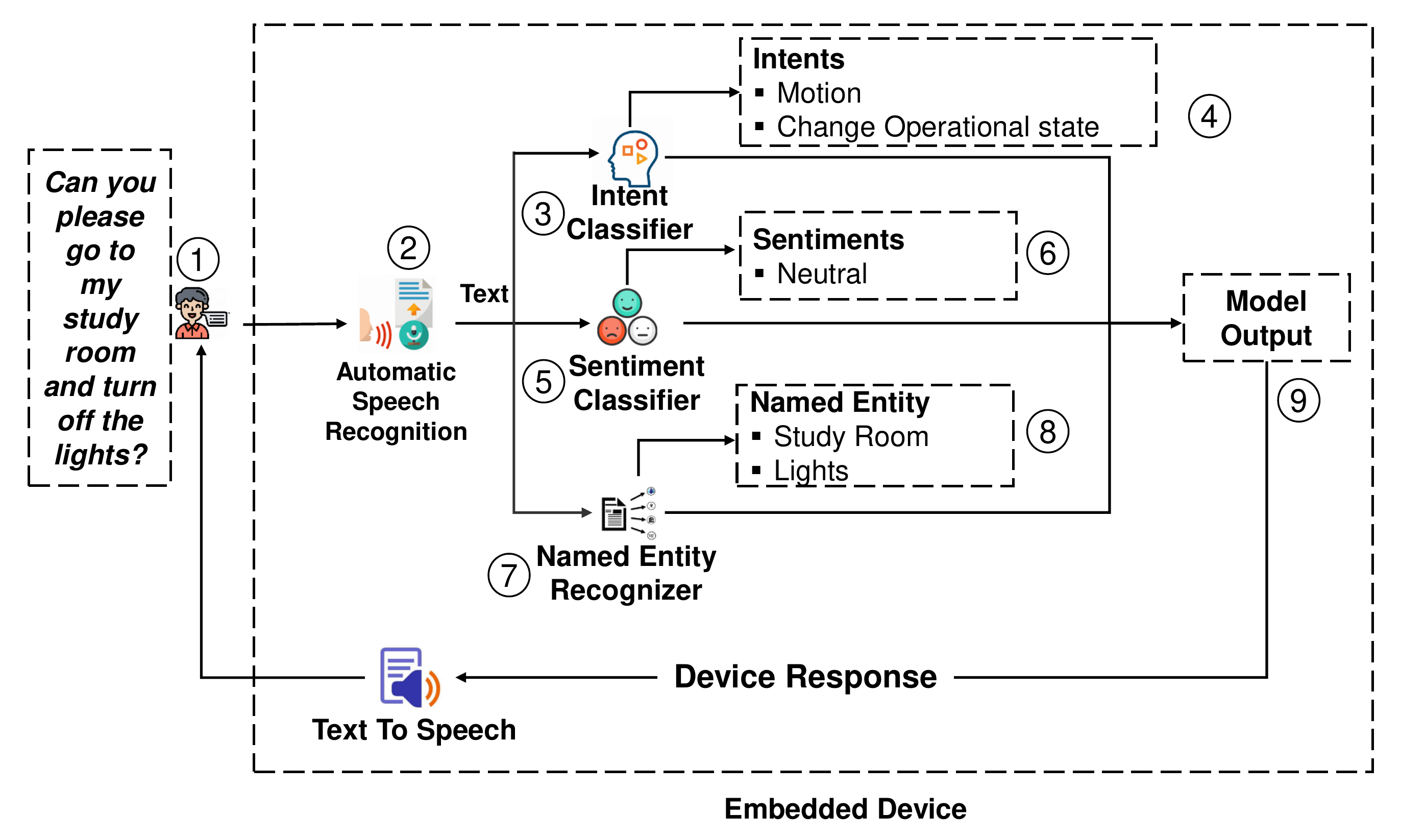}
        \caption{Utterance processing steps of a voice-controlled embedded device.}
        \label{fig:dialog_system}
\end{figure*}

\section{Problem Statement \& Datasets}\label{problem_statement}
\normalem


We aim to study \textit{how language models can be optimally deployed} to accomplish dialog processing in embedded devices.
The core technical challenge of any dialog system is to accurately understand and interpret user ``utterances'' and perform the right ``action'' accordingly. At a fundamental level, user utterance understanding relies on the following three basic NLP tasks\footnote{Section~\ref{ps:tasks} formally presents these three NLP tasks.}:~\ca \textit{Intent classification (IC)} --- to understand the need of the user, \cb \textit{Sentiment Classification (SC)} --- to understand user emotions, and \cc \emph{Named-entity Recognition (NER)} --- to extract related entities such as persons or objects. 

Figure~\ref{fig:dialog_system} presents the workflow of the dialogue-based systems considered in this work. The user initiates the interaction by providing a spoken command to the voice-controlled device (marker \textcircled{1} in Fig.~\ref{fig:dialog_system}). The device employs existing automatic speech recognition techniques~\cite{juang2005automatic,yu2016automatic} to convert the user's speech into texts as most language models take textual input (\textcircled{2}). The system then runs an ``intent classifier'' (Sec.~\ref{ps:intent}) and analyzes the extracted text to determine the user's \textit{intention} (\textcircled{3}). The classifier identifies the relevant user intentions for the given command (\textcircled{4}). For instance, the intents for the given command \textit{``Can you please go to my study room and turn off the lights?''} could be identified as \textit{``Motion''} and \textit{``Change operational state,''} as it instructed the device to move from its current position. Simultaneously, a ``sentiment classifier''(Sec.~\ref{ps:sentiment}) is employed to extract the user \textit{sentiment} (\textcircled{5}). In this case, the extracted sentiment is \textit{``Neutral''} as the command does not express any emotion (\textcircled{6}). Sentiment classification helps the voice-controlled device grasp the emotional context behind user utterances, enabling it to respond appropriately.  In addition, the dialog system utilizes a ``named entity recognizer'' (Sec.~\ref{ps:ner}) to identify specific ``entities'' (\textcircled{7}-\textcircled{8}). NER is crucial for accurately identifying user-specified entities like locations and objects, ensuring precise execution of commands in this scenario. For example, the entities, in this case, are \textit{``study room''} (location) and \textit{``lights''} (object). Once the user's intention and relevant entities are identified (\textcircled{9}), the control application running on the embedded device  carries out the specified task and sends a response back to the user. In this paper, we focus on understanding how the language models perform on embedded platforms for IC, SC, and NER tasks (\eg steps \textcircled{3}-\textcircled{8}).

\subsection{NLP Tasks under Consideration}\label{ps:tasks}



Recall from our earlier discussion that intent/sentiment classification and named-entity recognition are fundamental NLP tasks for any voice-controlled interactive system, chatbots, and virtual assistants. We now formally introduce IC, SC, and NER tasks.

\subsubsection{Intent Classification}\label{ps:intent}

To produce an accurate response, reduce backtracking, and minimize user frustration, Intent Classification (IC) is needed to identify which subsequent action a robot needs to perform depending on the user's utterance. A formal definition of IC can be given as follows: 
\begin{definition}Given a collection of user utterances $U = \{u_1, u_2, ..., u_n\}$, and a set of intent labels $I_x = \{i_1, i_2, ..., i_m\}$, classify each utterance $u_j$ $\epsilon$ $U$ with one to more intents labels from $I_x$.\end{definition}
Importantly, a user might have more than one intent while speaking to a robot and understanding the implicit or explicit intent expressed in a statement is essential to capturing the user's needs. For example, consider the following command:
~\enquote{{\centering\textit{Can you please go to my study room and turn off the lights?}}}
The command wants the robot to turn off the lights in the study room; the relevant intent here is \enquote{\textit{Change Operational State}}. However, the statement also expressed another intent related to \enquote{\textit{Motion}}, as the command requires the robot to change location. Without identifying all the underlying intents, the system cannot perform the right next step. Hence, recognizing and understanding \textit{all} types of intents stated in an utterance is crucial for accomplishing the eventual goal.

\subsubsection{Sentiment Classification} \label{ps:sentiment}


Sentiment analysis is regarded as an important task for accurate user modeling in natural dialog-based interactions, where user utterances are usually classified into multiple emotion/sentiment labels. A formal definition can be given as follows: 
\begin{definition}Given a collection of user utterances $U = \{u_1, u_2, ..., u_n\}$, and a set of sentiment labels $S_x = \{s_1, s_2, ..., s_m\}$, classify each expression $u_j$ $\epsilon$  $U$ with one to more sentiment labels from $S_x$.\end{definition}

For example, the following user utterance,\textit{~\enquote{OMG, yep!!! That is the final answer. Thank you so much!}} will be classified with sentiment labels~\enquote{gratitude} and,~\enquote{approval}. Similarly, statements such as\textit{~\enquote{This caught me off guard for real. I'm actually off my bed laughing}} will be labeled as~\enquote{surprise} and ~\enquote{amusement.}

\subsubsection{Named-entity Recognition} \label{ps:ner}

Named entity recognition (NER) --- often referred to as entity chunking, extraction, or identification --- is a sub-task of information extraction that seeks to locate and classify named entities mentioned in unstructured text. An entity can be expressed by a single word or a series of words that consistently refer to the same thing. Each detected entity is further classified into a predetermined category. The formal definition of the NER task can be given as follows:
\begin{definition}
Given a collection of statements/texts $S = \{s_1, s_2, ..., s_n\}$, and a set of entity labels $E_x = \{e_1, e_2, ..., e_m\}$, all the words/tokens in the text will be classified with an entity label $e_i$ $\epsilon$  $E_x$.
\end{definition}

NER can be framed as a sequence labelling task that is performed in two steps, first, detecting the entities from the text, and second, classifying them into different categories. A named entity recognizer model classifies each word/phrase representing an entity into one of the four types: \ca persons (PER), \cb objects (OBJ), \cc locations (LOC), and \cd miscellaneous names (MISC).

\subsection{Datasets}

Our study includes the following datasets\footnote{Appendix~\ref{huric_appen}-~\ref{wnut_appen} present additional details about these datasets.}: \ca HuRIC (for IC), \cb  GoEmotion (for SC), and \cc CoNLL and WNUT17 (for NER), as we present below. 

\subsubsection{Intent Classification: HuRIC}\label{intent_datatset}

For IC, we use Human Robot Interaction Corpus (HuRIC)~\cite{huric}, which is the state-of-the-art single-class classification dataset. The basic idea of HuRIC is to build a reusable corpus for human-robot interaction in natural language for a specific application domain, \ie house service robots. HuRIC includes a wide range of user utterances given to a robot representing different situations in a house environment. The motivation behind selecting HuRIC for our intent classification task stems from our specific interest in utilizing a dataset that captures human-robot conversations. The HuRIC dataset allows us to train and evaluate intent classification models on realistic dialogues between humans and robots in real-world scenarios. Table~\ref{table:huric} presents some statistics of HuRIC.



\begin{table}[!htb]
        \centering
        \begin{tabular}{r|c}
        \hline
        \bfseries Statistic & \bfseries Count\\
        \hline
        \hline
        Number of examples & 729 \\
        \hline
        Number of intent labels &  11 \\
        \hline
        Size of training dataset & 583 \\
        \hline
        Size of test dataset &  146 \\
        \hline
        \end{tabular}
        \vspace{1mm}
        \caption {Statistics of HuRIC dataset.}
        \label{table:huric}
\end{table}

\subsubsection{Sentiment Classification: GoEmotion}\label{subsec:sentiment-dataset}




We use GoEmotion~\cite{goemotions} dataset from Google AI for the SC task. GoEmotion is a human-annotated dataset of 58,000 Reddit comments extracted from popular English-language subreddits and labeled with 27 emotion categories. As the largest fully annotated English language fine-grained emotion dataset to date, the GoEmotion taxonomy was designed with both psychology and data applicability in mind Table~\ref{table:sentiment_dataset} presents some statistics of GoEmotion. We chose the GoEmotion dataset for SC task to ensure rigorous testing of our models. With its comprehensive emotion coverage and nuanced labeling, GoEmotion serves as a challenging yet realistic benchmark for evaluating sentiment detection performance.


\begin{table}[!htb]
        \centering
        \begin{tabular}{r|c}
        \hline
        \bfseries Statistic & \bfseries Count\\
        \hline
        \hline
        Number of labels & 27 + Neutral \\
        \hline
        Maximum sequence length in overall datasets & 30 \\
        \hline
        Size of training dataset & 43,410  \\
        \hline
        Size of test dataset &  5,427 \\
        \hline
        Size of validation dataset & 5,426\\
        \hline
        \end{tabular}
        \vspace{1mm}
        \caption {Statistics of GoEmotion dataset.}
        \label{table:sentiment_dataset}
    \end{table}


\subsubsection{Named-entity Recognition: CoNLL \& WNUT17}
\label{subsec:ner-dataset}

For NER we consider two datasets, \viz CoNLL~\cite{conll} and WNUT17 \cite{wnut17}.

\paragraph{CoNLL} CoNLL-2003 \cite{conll} was released as a part of CoNLL-2003 shared task: language-independent named entity recognition. The English corpus from this shared task consists of Reuters news stories between August 1996 and August 1997, each annotated with the entities associated with them. The data set consists of a training file, a development file, and a test file. The details of CoNLL-2003 are presented in Table~\ref{table:conll}.


\begin{table}[!t]
        \centering
        \begin{tabular}{r|c|c|c}
        \hline
        \bfseries Statistic & \bfseries Articles & \bfseries Sentences & \bfseries Tokens\\
        \hline
        \hline
        Training set & 946 & 14,987 & 203,621\\
        \hline
        Development set & 216 &  3,466 & 51,362 \\
        \hline
        Test set & 231 & 3,684 & 46,435  \\
        \hline
        
        \end{tabular}
        \vspace{1mm}
        \caption {Statistics of CoNLL dataset.}
        \label{table:conll}
    \end{table}

\smallskip

\paragraph{WNUT17.} 
While the CoNLL corpus is based on news stories, we wanted to select a dataset that contains user utterances such as those available on HuRIC. Unfortunately, we could not find such a NER dataset but discovered a very similar corpus (WNUT2017~\cite{wnut17}) that contains user-generated text. The WNUT2017 dataset's shared task focuses on identifying unusual, previously-unseen entities in the context of emerging discussions. Identifying entities in noisy text is really challenging, even for human annotators, due to novel entities and surface forms. In this dataset, user comments were mined from different social media platforms because they are large, and samples can be mined along different dimensions, such as texts from/about geo-specific areas, about home aid, and particular topics and events. Table \ref{table:ner_dataset} summarizes the dataset properties.

\begin{table}[!htb]
        \centering
        \begin{tabular}{r|c}
        \hline
        \bfseries Statistic & \bfseries Count\\
        \hline
        \hline
        Number of examples & 5690 \\
        \hline
        Number of labels & 6 \\
        \hline
        Size of training dataset & 3394  \\
        \hline
        Size of test dataset &  1287 \\
        \hline
        Size of validation dataset & 1009\\
        \hline
        \end{tabular}
        \vspace{1mm}
        \caption {Statistics of WNUT17 dataset.}
        \label{table:ner_dataset}
    \end{table}
    
\section{Experimental Setup}\label{exp_design}
We now summarize BERT architectures and configurations used in our experiments (Sec.~\ref{subsec:methods} and Sec.~\ref{pruning}). We selected four off-the-shelf platforms from different chip vendors (Intel, AMD, ARM, NVIDIA) to understand the feasibility of using NLP tasks on a variety of architectures (Sec.~\ref{sec:embd_plat}). To measure the performance of each task (IC, SC, NER), we use three popular metrics (\ie Precision, Recall, and $F_1$ score). Section~\ref{sec:rq} lists the design questions explored in our investigation. 
The blueprints of our implementation, including related code/documentation, is \textbf{publicly available} for community use~\cite{repo_link}. 


\begin{table}[!t]
  \centering
  \resizebox{1.01\columnwidth}{!}{
  \begin{tabular}{@{}l|c|c|c|c@{}}
    \hline
    \bfseries  &
    \bfseries BERT   & 
    \bfseries RoBERTa & 
    \bfseries DistilBERT & 
    \bfseries TinyBERT\\ \hline\hline
    Number of Layers  & 12 & 12 & 6 & 4  \\
    Attention Heads  & 12  & 12 & 12  & 12 \\
    Hidden Layer Size  & 768 & 768  & 768 & 312 \\
   Feed-Forward Layer Size & 3072 & 3072 & 3072 & 1200 \\
   Vocabulary Size & 30522 & 30522 & 30522 & 30522 \\ \hline
  \end{tabular}
  }
  \vspace{1mm}
  \caption{Attributes of the BERT variants used in our study.}
  \label{table:nlp_models}
\end{table}

\subsection{Off-the-Shelf BERT Variants}
\label{subsec:methods}


We use a pre-trained base variant of BERT \cite{bertDelvin}, RoBERTa \cite{Roberta}, DistilBERT \cite{distilbert}, TinyBERT \cite{tinybert} model from Huggingface\footnote{https://huggingface.co/.} and finetune the models on respective datasets (\ie HuRIC, GoEmotion, CoNLL, and WNUT17). 
Table~\ref{table:nlp_models} listed the parameters of the BERT variants and Table~\ref{table:hp-ner} presents the hyper-parameter used in our experiments.

\begin{table}[!htb]
    \centering
    \begin{tabular}{l|c|c}
    
         \hline 
         \bfseries Hyperparameter   & \bfseries NER & \bfseries IC/SC \\\hline\hline
         Number of epochs & 3 & 3\\
         Batch size & 64 & 64\\
         Learning rate & $[e^{-6}, e^{-4}]$ & $[e^{-6}, e^{-4}]$\\
         Weight decay & $[0.01, 0.3]$ & $[0.01, 0.3]$\\
         Optimizer & Adam & Adam\\
         Adam epsilon & $1e^{-8}$ & $1e^{-8}$\\
         Max sequence length & 64 & 128\\
         \hline
    \end{tabular}
    \vspace{1mm}
    \caption{Hyperparameter values for finetuning BERT on IC/SC and NER tasks.}
    \label{table:hp-ner}
\end{table}

\begin{table*}[!t]
        \centering
        \begin{tabular}{cc||c|c|c|c}
        \hline
        \multicolumn{2}{c||}{\bfseries Embedded Platform} & \bfseries Architecture & \bfseries CPU & \bfseries GPU & \bfseries Memory\\
    	\hline
        \hline
        \adjustbox{valign=c}{\includegraphics[scale=0.06]{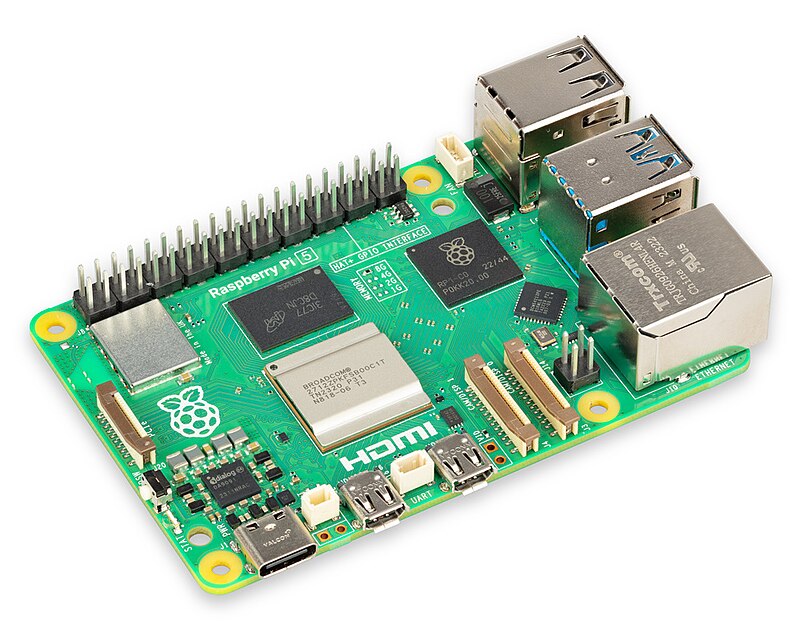}} & \rpi   & ARM &  Quad-core Cortex-A72 & \faClose  & 2 GB and 4 GB  \\
        \hline
        \adjustbox{valign=c}{\includegraphics[scale=0.055]{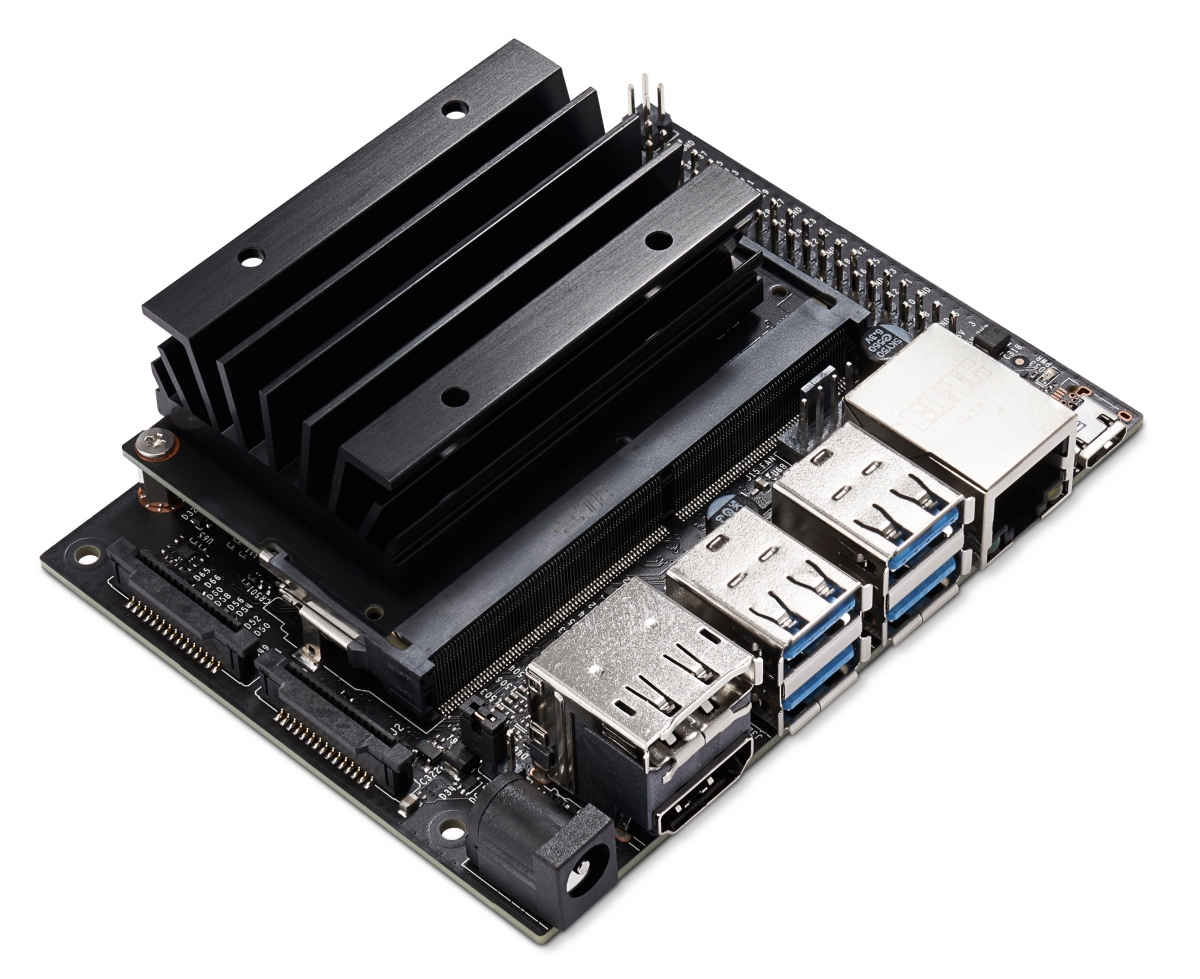}} & \jetn  & ARM & Quad-core Cortex-A57  & 128-core NVIDIA Maxwell & 2 GB and 4 GB  \\
        \hline
        \adjustbox{valign=c}{\includegraphics[scale=0.06]{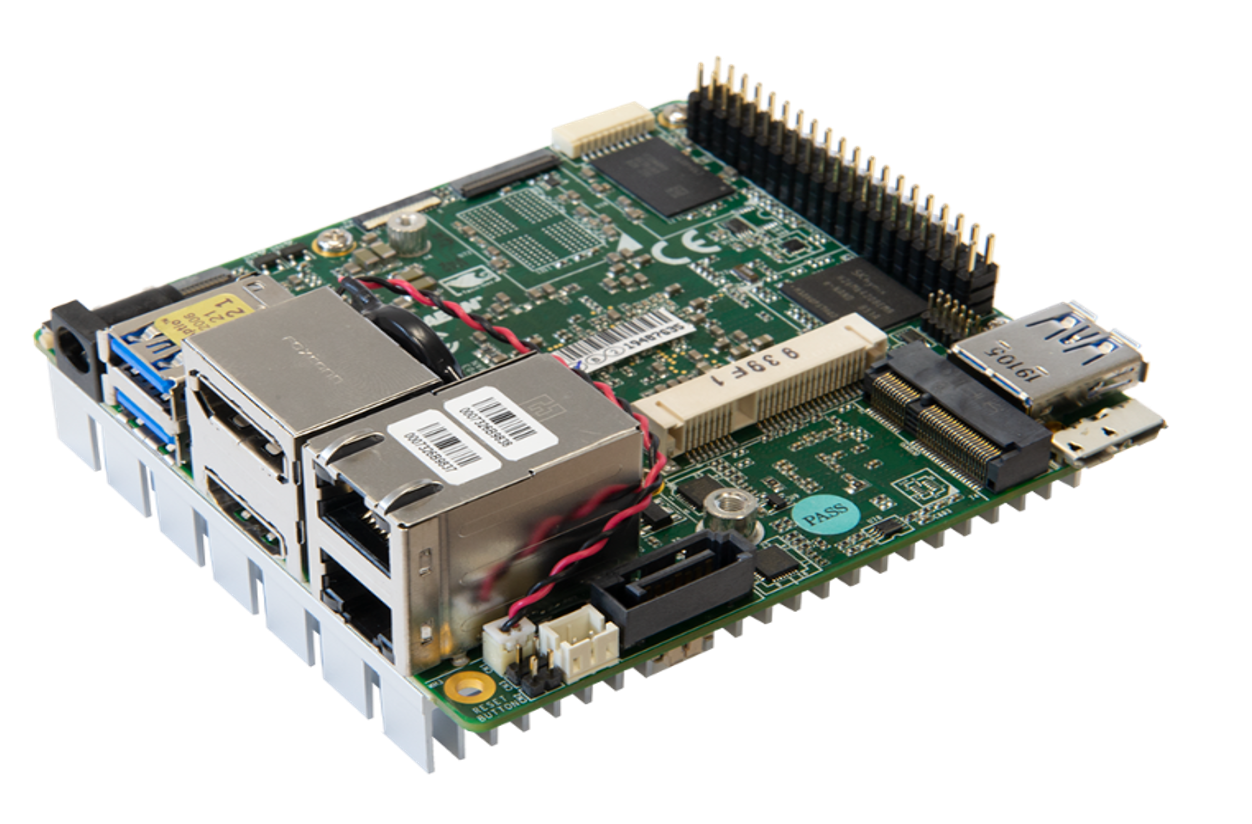}} & \up   & x86  &  Dual-core Intel Celeron N6210 & \faClose & 2 GB and 4 GB  \\
        \hline
        \adjustbox{valign=c}{\includegraphics[scale=0.09]{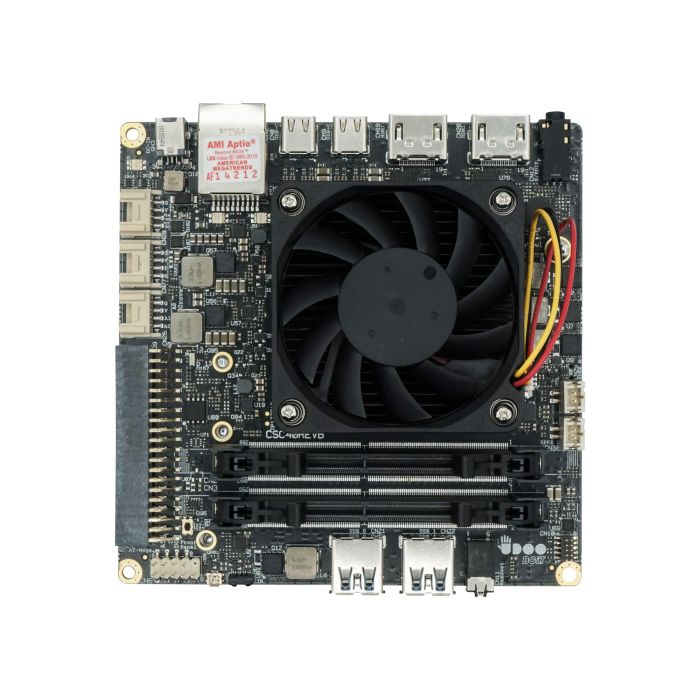}} & \udob & x86  &  Quad-core AMD Ryzen V1605B  &  \faClose &  2 GB and 4 GB \\
        \hline
        \end{tabular}
        \vspace{1mm}
        \caption{Embedded platforms used in our evaluations: \ca~\rpi~\cite{raspberrypi}, \cb~\jetn~\cite{jetsonnano}, \cc~\up~\cite{upboard}, and \cd~\udob~\cite{udoobolt}. We used 2 GB and 4 GB memory configurations for each board.}
        \label{table:hw_config}
\end{table*}


\subsection{Pruning and Custom Configurations} \label{pruning}

We also experiment with custom, smaller BERT configurations. Due to the resource constraints (\eg memory and energy limits) of embedded devices, it is necessary to explore different variants of BERT-based models that can be optimized to run on the device. We can reduce the model size on two fronts: \ca by reducing the layer size and \cb  by pruning various attributes. 


In our study, we experiment with~\emph{different layer combinations} of BERT models and test their performance on~\emph{different hardware configurations}, which are presented in Table~\ref{Performancetable_sentiment1}, and \ref{Performancetable_NER1}. With two layers of $BERT_{Base}$ (instead of 12), the model size reduces significantly, but so does the accuracy (in terms of $F_1$ score). Still, these models give better accuracy than the distilled models on complex NLP tasks. Also, where $BERT_{Base}$ model with 12 layers cannot run on a resource-constrained device, using a lesser number of layers enable a model to execute on tiny devices with good accuracy compared to the distilled methods (DistilBERT and TinyBERT).

For further shrinking of the model, pruning can be applied to weights, neurons, layers, channels, and attention heads, depending on the heuristic used. In this paper, we focus on pruning  \emph{attention heads}, which plays an important role in the model's performance and contributes a large number of parameters. Although multi-headed attention is a driving force behind many recent state-of-the-art models, \citet{sixteen_heads} finds that even if models have been trained using multiple heads, in practice, a large percentage of attention heads can be removed without significantly impacting performance. In fact, in some layers, \emph{attention heads} can even be reduced to a single head. 

Based on this fact, we experiment with reducing the size of $BERT_{Base}$ by dynamically pruning the \emph{attention heads}. 
Each attention head provides a distribution of attention for the given input sequence. For each attention head, we calculate the head importance by calculating the \textit{entropy} of each head.
After that, we mask all the attention heads with the lowest entropy. This process is repeated for each layer of the encoder.
After masking the heads, we calculate the overall ${F_1}$ score of the masked model and determine the drop in ${F_1}$ score compared to the original unpruned model. If the drop is less than a predefined threshold, we prune the masked attention heads. We repeat the process until the drop in ${F_1}$ score reaches the predefined threshold.  
This pruning procedure reduces the model size significantly while maintaining the desired model performance. 
%
%
%
%
%
%
%

\subsection{Evaluation Platforms} \label{sec:embd_plat}




We evaluate the BERT models on heterogeneous setup, \viz  x86 (Intel and AMD) and ARM platforms, including devices with or without GPU.
In particular, we used the following
four different embedded platforms: \ca~\rpib~\cite{raspberrypi}, \cb~\jetn~\cite{jetsonnano}, \cc~\up~\cite{upboard}, and \cd~\udob~\cite{udoobolt}. Table~\ref{table:hw_config} lists the hardware configurations used in our setup. Among them, only \jet board equipped with GPU (128 NVIDIA CUDA cores). We used 2 GB and 4 GB memory configurations for each of the four boards. The SoC (System on Chip) of ARM-based boards (\rpi and \jet) configured with soldered memory --- hence, we used two different boards for each with 2 GB and 4 GB RAM configurations. The x86-based boards (\ie \up and \udo) are modular, so we used the same board but two different DDR4 RAMs (2 GB and 4 GB). Hence, our evaluation setup consists of \textit{eight different hardware configurations} running on four ARM and two x86 boards. For energy measurements during the inference steps, we used UM25C energy meter~\cite{um25c}. The total cost for our test harness is \$1697 (see Table~\ref{tab:board-costs-updated} in Appendix~\ref{estimated_cost} for a detailed breakdown). We performed all experiments on Linux kernel 5.15.0. The NLP models were developed using PyTorch library (version 1.13).

\subsection{Design Challenges \& Research Questions} \label{sec:rq}

We conducted extensive experiments to investigate the following research questions (RQs).
\begin{itemize}[itemsep=0.2ex,partopsep=0ex,parsep=0ex]
	
    \item \textbf{RQ~1.} Given specific user-defined constraints, such as system resources (processor, memory) and performance budgets (accuracy, inference time), what is the optimal (if any) BERT-based architecture satisfying those constraints?
	
    \item \textbf{RQ~2.} What is the accuracy vs. model-size trade-off as we prune the models?
	
    \item  \textbf{RQ~3.} What are the trade-offs of accuracy and corresponding resource usage (\eg memory, inference-time, energy consumptions) as we perform pruning?

    \item  \textbf{RQ~4.} {Does GPU aid in inference time?}
    
     \item  \textbf{RQ~5.} {What are the energy consumption differences among various architectures (x86 and ARM)? Does the presence of GPUs impact energy usage?}


\end{itemize}



%








\section{Results}\label{results}
In this section, we report our results for the three basic NLP tasks, \ie IC, SC, and NER for both existing (\eg BERT, RoBERTa, DistilBERT, and TinyBERT) and custom BERT architectures.

\begin{table*}[!htb]
\centering
\resizebox{0.9\textwidth}{!}{%
\begin{tabular}{  c c c |  c c c | c c c |  c c c}
\hline 
\multicolumn{12}{c}{\textbf{Intent Classification Task (Dataset: HuRIC)}}\\
\hline 
\hline 
\multicolumn{3}{c|}{\textbf{BERT}} & \multicolumn{3}{c|}{\textbf{RoBERTa}} & \multicolumn{3}{c|}{\textbf{DistilBERT}} & \multicolumn{3}{c}{\textbf{TinyBERT}} \\
\hline
\textbf{Precision} & \textbf{Recall} & $\boldsymbol{F_1}$ & \textbf{Precision} & \textbf{Recall} & $\boldsymbol{F_1}$ & \textbf{Precision} & \textbf{Recall} & $\boldsymbol{F_1}$ & \textbf{Precision} & \textbf{Recall} & $\boldsymbol{F_1}$
\\ \hline
0.943 & 0.985 & 0.961 & 0.975 & 0.952 & 0.962 & 0.951 & 0.903 & 0.927 & 0.912 & 0.897 & 0.902 \\
\hline
\end{tabular}%
}
\vspace{1mm}
\caption{\label{IntentPerformance} Performance of BERT, RoBERTa, DistilBERT, and TinyBERT for the IC task.}
\vspace{-4mm}
\end{table*}

\begin{table*}[!htb]
\centering
\resizebox{0.9\textwidth}{!}{%
\begin{tabular}{  c c c |  c c c | c c c |  c c c}
\hline 
\multicolumn{12}{c}{\textbf{Multi-label Sentiment Detection Task (Dataset: GoEmotion)}}\\
\hline 
\hline 
\multicolumn{3}{c|}{\textbf{BERT}} & 
\multicolumn{3}{c|}{\textbf{RoBERTa}} & 
\multicolumn{3}{c|}{\textbf{DistilBERT}} & \multicolumn{3}{c}{\textbf{TinyBERT}} \\
\hline
\textbf{Precision} & \textbf{Recall} & $\boldsymbol{F_1}$ & \textbf{Precision} & \textbf{Recall} & $\boldsymbol{F_1}$ & \textbf{Precision} & \textbf{Recall} & $\boldsymbol{F_1}$ & \textbf{Precision} & \textbf{Recall} & $\boldsymbol{F_1}$
\\ \hline
0.77 & 0.37 & 0.490 & 0.731 & 0.452 & 0.567 & 0.174 & 0.121 & 	0.134 & 0.060 & 0.030 & 0.031\\
\hline

\end{tabular}%

}
\vspace{1mm}
\caption{\label{SCPerformance} Performance of BERT, RoBERTa, DistilBERT, and TinyBERT for the SC task.}

\end{table*}

\begin{table*}[!htb]
\centering
\resizebox{0.9\textwidth}{!}{%
\begin{tabular}{  c c c |  c c c | c c c |  c c c}
\hline 
\multicolumn{12}{c}{\textbf{Named-entity Recognition Task}}\\
\hline 
\hline
\multicolumn{12}{c}{\textbf{Dataset: CoNLL}}\\
\hline 
\multicolumn{3}{c|}{\textbf{BERT}} & \multicolumn{3}{c|}{\textbf{RoBERTa}} & \multicolumn{3}{c|}{\textbf{DistilBERT}} & \multicolumn{3}{c}{\textbf{TinyBERT}} \\
\hline
\textbf{Precision} & \textbf{Recall} & $\boldsymbol{F_1}$ & \textbf{Precision} & \textbf{Recall} & $\boldsymbol{F_1}$ & \textbf{Precision} & \textbf{Recall} & $\boldsymbol{F_1}$ & \textbf{Precision} & \textbf{Recall} & $\boldsymbol{F_1}$
\\ \hline
0.891 & 0.963 & 0.926 & 0.882 & 0.955 & 0.917 & 0.906 & 0.967 & 0.935 & 0.872 & 0.958 & 0.911\\
\hline\hline
\multicolumn{12}{c}{\textbf{Dataset: WNUT17}}\\
\hline 
\multicolumn{3}{c|}{\textbf{BERT}} & \multicolumn{3}{c|}{\textbf{RoBERTa}} & \multicolumn{3}{c|}{\textbf{DistilBERT}} & \multicolumn{3}{c}{\textbf{TinyBERT}} \\
\hline
\textbf{Precision} & \textbf{Recall} & $\boldsymbol{F_1}$ & \textbf{Precision} & \textbf{Recall} & $\boldsymbol{F_1}$ & \textbf{Precision} & \textbf{Recall} & $\boldsymbol{F_1}$ & \textbf{Precision} & \textbf{Recall} & $\boldsymbol{F_1}$
\\ \hline
 0.671 & 0.295 & 0.410 & 0.537 & 0.315 & 0.397 & 0.316 & 0.014 & 0.028 & 0.0 & 0.0 & 0.0\\
\hline

\end{tabular}%
}
\vspace{1mm}
\caption{\label{NERPerformance} Performance of BERT, RoBERTa, DistilBERT, and TinyBERT for the NER task.}

\end{table*}

\subsection{Experience with Existing BERT Variants}

\paragraph{Intent Classification (IC)} 
Recall from our earlier discussion (Sec.~\ref{intent_datatset}) that we used the HuRIC dataset for IC. Table~\ref{IntentPerformance} presents our findings after running IC tasks using the HuRIC dataset on different hardware. We observed that all the models performed similarly on this dataset, achieving more than 90\% $F_1$ Score.

\paragraph{Multi-label Sentiment Classification (SC)} We next analyze the performance of multi-label SC tasks on \hardware board. As mentioned in Sec.~\ref{subsec:sentiment-dataset}, we used GoEmotion~\cite{goemotions} dataset for this task. GoEmotion includes direct user-to-user conversation text and labels them with many user emotions. Table~\ref{SCPerformance} summarizes the performance of all the models for this task. Interestingly, for this task, DistilBERT and TinyBERT failed drastically, as they achieved a very low $F_1$ Score. The failure of distilled models can be attributed to the difficulty of the task. Multi-label SC requires each utterance to be classified with more than one sentiment. Therefore, this dataset is not a straightforward positive-negative sentiment detection.  

\paragraph{Named-entity Recognition (NER)} As we mention in Sec.~\ref{subsec:ner-dataset}, we use two different datasets to test the NER task. Table~\ref{NERPerformance} summarizes the performance over both the NER datasets and shows that for the CoNLL dataset. In this setup, all models performed comparatively the same. However, the performance of distilled models dropped sharply for the WNUT17 dataset (which focuses on identifying unusual, previously-unseen entities). This drop tends to be due to the difficulty of analyzing this task, as NER evaluates the ability to detect and classify novel, emerging, singleton-named entities in noisy inputs.



In summary, our findings are as follows.

{\centering
\begin{tcolorbox}[colback=gray!20,colframe=black, boxrule=0.0pt, width=\columnwidth, left=0.5pt,
	right=0.5pt,
	top=0.5pt,
	bottom=0.5pt]
\begin{itemize}[leftmargin=*]
    \item All models achieved decent F1 scores (>90\%) for IC task.
    \item DistilBERT and TinyBERT struggled with the multi-label SC task as none achieved an F1 score of more than 15\%.
    \item All BERT models excelled in the NER task on the CoNLL dataset and accurately recognized named entities (resulting in >90\% F1 scores).
    \item Distilled models showed a performance decline for the NER task on the WNUT17 dataset with the F1 score dropping to less than 5\%, indicating difficulty with dataset intricacies.
\end{itemize}
\end{tcolorbox} %
}


\begin{table*}[!htb]\huge
    \centering
   \begin{adjustbox}{width=\linewidth,center}%
    \begin{tabular}{p{1.5cm}|c|c|c|c|c|c}
    \hline 
    \multirow{2}{*}{\textbf{Task}} & \multirow{2}{*}{\textbf{Metrics}}
    &  \multicolumn{5}{c}{\bf $\boldsymbol F_1$ Score Threshold ($\boldsymbol \theta$)} \\ \cline{3-7} & 
    & { $\boldsymbol\theta_{50}$} 
    & {$\boldsymbol\theta_{60}$} 
    & {$\boldsymbol\theta_{70}$} 
    & {{$\theta_{80}$}} 
    & {\textbf{$\theta_{90}$}} \\\hline\hline

    \hline
   IC   
            & \begin{tabular}{@{}c@{}}Layer\\
                                     MS\\ 
                                     Params\\
                                     Pruning\\ 
            \end{tabular}
        & \begin{tabular}{p{1.2cm}|p{1.1cm}|p{1.0cm}|p{1.1cm}}
         2&4&6&8\\\hline
         144.3 & N/A & N/A & N/A \\
         35.1 &  & & \\
         68\% &  &  & \\
        \end{tabular}

        &~\begin{tabular}{p{1.1cm}|p{1.1cm}|p{1.0cm}|p{1.1cm}}
         2&4&6&8\\\hline
         148.3 & N/A  &N/A & N/A \\
         37.1 &  & & \\
         66\% &  & & \\
        \end{tabular}

        & \begin{tabular}{p{1.0cm}|p{1.2cm}|p{1.1cm}|p{1.1cm}}
         2&4&6&8\\\hline
         N/A & 195.6  &N/A & 282.2 \\
         & 48.9   & &70.5 \\
         &  56\% &   & 36\% \\
        \end{tabular} 
                            
        & \begin{tabular}{p{1.1cm}|p{1.1cm}|p{1.1cm}|p{1.1cm}}
         2&4&6&8\\\hline
         154.2 & 198.7  & 246 & N/A \\
         38.6 & 49.7 & 61.5 & \\
         65\% & 55\% & 44\% & \\
        \end{tabular}

        & \begin{tabular}{p{1.1cm}|p{1.1cm}|p{1.1cm}|p{1.1cm}}
                                     2&4&6&8\\\hline
                                     N/A & 211.3& 268 & 303.5 \\
                                      & 52.8 & 67 & 75.9 \\
                                     & 52\% & 39\% & 31\% \\
        \end{tabular} \\ 
    \hline

 NER      
        & \begin{tabular}[c]{@{}c@{}}Layer\\
                                     MS\\ 
                                     Params\\
                                     Pruning\\

        \end{tabular}

        & \begin{tabular}{p{1.2cm}|p{1.1cm}|p{1.0cm}|p{1.1cm}}
                                    2 & 4 & 6 & 8\\\hline
                                     136.4 & N/A & N/A & N/A\\
                                     34.1 &  & &\\
                                     67\% & & &\\
   
        \end{tabular}   
        & \begin{tabular}{p{1.2cm}|p{1.1cm}|p{1.0cm}|p{1.1cm}}
                                    2 & 4 & 6 & 8\\\hline
                                     147.4 & N/A & N/A & N/A\\
                                     36.9 &  & &\\
                                     65\% & & &\\

        \end{tabular}   
        & \begin{tabular}{p{1.1cm}|p{1.1cm}|p{1.1cm}|p{1.1cm}}
                                    2 & 4 & 6 & 8\\\hline
                                     N/A & 185.2 & 233.3 & N/A\\
                                      & 46.3 & 58.3 &\\
                                      & 57\% & 45\% &\\
               
        \end{tabular}                        
        & \begin{tabular}{p{1.1cm}|p{1.1cm}|p{1.1cm}|p{1.1cm}}
                                    2 & 4 & 6 & 8\\\hline
                                      N/A & 201 & 249.8 & 268\\
                                     & 46.3 & 62.4 & 67\\
                                      & 45\% & 57\% & 38\%\\

        \end{tabular}                        
        
        & \begin{tabular}{p{1.1cm}|p{1.1cm}|p{1.1cm}|p{1.1cm}}
                                    2 & 4 & 6 & 8\\\hline
                                      N/A & N/A & 260.8 & 289.2\\
                                      &  & 65.2 & 72.3\\
                                      & & 39\% & 32\%\\
   
        \end{tabular}   \\

    \hline
                                                   
    \hline

    \end{tabular}
 \end{adjustbox}
 \vspace{1mm}
 \caption{\label{Performancetable_task} 
 Performance of SC and NER tasks for the GoEmotion and WNUT17 datasets on various configurations.~In metrics column, MS= Model Size (MB), Params= Parameters (Million).
 }
 \label{intentparamstable}
\end{table*}

\subsection{Exploration with Custom Architectures}\label{design}

 Based on our experiment results (Tables~\ref{IntentPerformance}--\ref{NERPerformance}), we further explore different alternative BERT-based architectures by reducing the layers and pruning the attention heads from the original $BERT_{Base}$ model. For this exploration, we primarily focus on the challenging tasks, \ie multi-label SC and NER where off-the-shelf models (\eg DistilBERT and TinyBERT) failed to perform.
 
 
 Table~\ref{Performancetable_task}--Table~\ref{Performancetable_ner} 
 present our exploration findings.\footnote{\textbf{\uline{Note:}} We omit the results for IC on custom BERT architectures as existing models suffice to perform this task.} We discuss our observations in Sec.~\ref{sec:observations} and provide answers to the research questions posed in Sec.~\ref{sec:rq}. Before we proceed with the discussion, we present a brief overview of the attributes and metrics used in our evaluation.
  
 
 \subsubsection{Model Attributes} In the evaluation, we vary the following model attributes.
 
\begin{itemize}[leftmargin=*,itemsep=0.2ex,partopsep=0ex,parsep=0ex]

\item ${F_1}$ \textit{Threshold} ($\theta$): The $\theta$-cells represents what percentage of the ${F_1}$ score (with respect to $BERT_{Base}$) is retained by the models. In our experiment, we varied $\theta$ between 50\% to 90\% and reported the model details in respective columns. For example, $\theta$ set to 80\% implies the $\theta_{80}$ column. 

\item \textit{Platform}: Indicates the memory capacity of the different hardware we used in our exploration. 

\item \textit{Layer}: Represents the number of layers retained.

\item \textit{Model Size}: The size of the saved model after training. We stored the saved model on the disk which is then loaded on the memory for inference. 
 
\item \textit{Parameters}: This metric indicates the total number of parameters in the saved model. We obtained the model parameters using the \texttt{model.parameters()} method in PyTorch~\cite{paszke2017automatic}.

\item \textit{Pruning}: Pruning percentage represents the reduction in size from the $BERT_{Base}$ model. For example, a pruning percentage of 70\% implies that the pruned model is 70\% smaller than $BERT_{Base}$.

\end{itemize}

\subsubsection{Performance Metrics} We consider the three metrics to benchmark the NLP models: \ca inference time, \cb memory usage, and \cc energy consumption, as we present below. 

\begin{itemize}[leftmargin=*,itemsep=0.2ex,partopsep=0ex,parsep=0ex]

\item \textit{Memory Consumption}: Maximum memory usage (in megabytes) of the corresponding NLP task running on different boards during the inference time. We used Python \texttt{memory\_profiler} for each input to get the memory usage.

\item \textit{Inference Time}: Depending on the specific task, the 95th percentile time required for the model to infer the appropriate \textit{Intent, Sentiment or Entity} from a given command.

\item \textit{Energy Consumption}: The average energy consumed (in watts) by the different hardware during the inference of a given command. We measured both the \textit{rest time} (\ie when the system is idle) and \textit{inference time} energy consumption.

\end{itemize}

\begin{table*}[!htb]\huge
    \centering
   \begin{adjustbox}{width=\linewidth,center}%
    \begin{tabular}{p{1.5cm}|c|c|c|c|c|c}
    \hline 
    \multirow{2}{*}{\textbf{Plat.}} & \multirow{2}{*}{\textbf{Metrics}}
    &  \multicolumn{5}{c}{\bf $\boldsymbol F_1$ Score Threshold ($\boldsymbol \theta$)} \\ \cline{3-7} & 
    & { $\boldsymbol\theta_{50}$} 
    & {$\boldsymbol\theta_{60}$} 
    & {$\boldsymbol\theta_{70}$} 
    & {{$\theta_{80}$}} 
    & {\textbf{$\theta_{90}$}} \\\hline\hline

    \hline
  Pi \par 2 GB   
            &\begin{tabular}{@{}c@{}}Layer\\
                                     EC\\ 
                                     MC\\ 
                                IT\\ 
            \end{tabular}
        &\begin{tabular}{p{1.1cm}|p{1.1cm}|p{1.1cm}|p{1.1cm}}
         \clrgray 2&4&6&8\\\hline
          \clrgray 4.24& N/A  & N/A & N/A \\
          \clrgray394.3&  &  & \\
         \clrgray0.52 &  &  & \\
        \end{tabular}
                                       
        &\begin{tabular}{p{1.1cm}|p{1.1cm}|p{1.1cm}|p{1.1cm}}
         \clrgray 2&4&6&8\\\hline
         \clrgray 4.28 & N/A  & N/A & N/A \\
          \clrgray 391.5& & & \\
          \clrgray 0.51&  & & \\
        \end{tabular}

        & \begin{tabular}{p{1.1cm}|p{1.1cm}|p{1.1cm}|p{1.1cm}}
         2&\clrgray4&6&8\\\hline
         N/A & \clrgray 4.32 & N/A  & 4.37  \\
         & \clrgray 440.4 & & 563 \\
         & \clrgray 1.1 &  & 2.17 \\
        \end{tabular} 
                            
        & \begin{tabular}{p{1.1cm}|p{1.1cm}|p{1.1cm}|p{1.1cm}}
         \clrgray2&4&6&8\\\hline
         \clrgray4.35& 4.3 & 4.48 &N/A \\
         \clrgray401.9 & 446.2  & 497.3 & \\
         \clrgray0.54 & 1.01  & 1.69 & \\
        \end{tabular}

        & \begin{tabular}{p{1.1cm}|p{1.1cm}|p{1.1cm}|p{1.1cm}}
                                     2&\clrgray4&6&8\\\hline
                                     N/A&\clrgray 3.91 & 4.45 & 4.34 \\ 
                              & \clrgray462.9  & 521.5 & 559.4 \\ 
                              &\clrgray 1.3 & 1.81 & 2.35 \\  
        \end{tabular} \\ 
    \hline

  Pi\par  4 GB      
        & \begin{tabular}[c]{@{}c@{}}Layer\\
                                     EC\\ 
                                     MC\\ 
                                IT\\

        \end{tabular}

        & \begin{tabular}{p{1.1cm}|p{1.1cm}|p{1.1cm}|p{1.1cm}}
         \clrgray2&4&6&8\\\hline
          \clrgray4.91& N/A  & N/A & N/A \\
          \clrgray397.5&  & & \\
         \clrgray0.44&  & &  \\
        \end{tabular}

        & \begin{tabular}{p{1.1cm}|p{1.1cm}|p{1.1cm}|p{1.1cm}}
         \clrgray2&4&6&8\\\hline
          \clrgray5.05& N/A & N/A & N/A\\
          \clrgray401&  & &\\
          \clrgray0.44&  & &  \\
        \end{tabular}
    
        & \begin{tabular}{p{1.1cm}|p{1.1cm}|p{1.1cm}|p{1.1cm}}
         2&\clrgray4&6&8\\\hline
          N/A& \clrgray4.67 &N/A & 5.09 \\
         & \clrgray446.3 & & 569.2\\
           & \clrgray0.82 & & 1.5 \\
        \end{tabular}
                               
        & \begin{tabular}{p{1.1cm}|p{1.1cm}|p{1.1cm}|p{1.1cm}}
                                    \clrgray2 & 4 & 6 & 8\\\hline
                                     \clrgray4.98 & 4.72 & 4.8 & N/A\\ 
                                  \clrgray404.2& 453.2 & 507.7 &\\ 
                                 \clrgray0.5& 0.81 & 1.21 &\\ 
        \end{tabular}                         
        
        & \begin{tabular}{p{1.1cm}|p{1.1cm}|p{1.1cm}|p{1.1cm}}
                                  N/A & \clrgray4 & 6 & 8\\\hline
                                 &\clrgray4.77  & 4.78 & 4.95 \\ 
                                & \clrgray464.6 & 528.8 & 563.4\\ 
                                & \clrgray1.00 & 1.46 & 1.55\\
        \end{tabular}  \\

    \hline

\hline
  Jetson 2 GB   
            & \begin{tabular}{@{}c@{}}Layer\\
                                     EC\\ 
                                     MC\\ 
                                IT\\ 
            \end{tabular}
        & \begin{tabular}{p{1.1cm}|p{1.1cm}|p{1.1cm}|p{1.1cm}}
        \clrgray 2&4&6&8\\\hline
          \clrgray 5.87&N/A  &N/A & N/A\\
          \clrgray299.2&  &  & \\
          \clrgray 0.29&  &  & \\
        \end{tabular}
                               
        & \begin{tabular}{p{1.1cm}|p{1.1cm}|p{1.1cm}|p{1.1cm}}
         \clrgray2&4&6&8\\\hline
          \clrgray5.96& N/A & N/A & N/A\\
          \clrgray320.1& & & \\
          \clrgray0.26&  & & \\
        \end{tabular}

        & \begin{tabular}{p{1.1cm}|p{1.1cm}|p{1.1cm}|p{1.1cm}}
         2&\clrgray4&6&8\\\hline
         N/A & \clrgray6.01  &  N/A & 6.28  \\
         & \clrgray345.4 & & 478.5 \\
         & \clrgray0.49 &  & 0.88 \\
        \end{tabular} 
                            
        & \begin{tabular}{p{1.1cm}|p{1.1cm}|p{1.1cm}|p{1.1cm}}
         \clrgray2&4&6&8\\\hline
        \clrgray5.87 & 6.14 & 6.26 & N/A \\
         \clrgray329.7 & 352.6 & 415 & \\
         \clrgray0.30 & 0.51 & 0.68 & \\
        \end{tabular}

        & \begin{tabular}{p{1.1cm}|p{1.1cm}|p{1.1cm}|p{1.1cm}}
                                     2&\clrgray4&6&8\\\hline
                                    N/A &\clrgray 6.13 & 6.22 &  6.24\\ 
                              & \clrgray367.5 & 429.8 & 463.9 \\ 
                              &\clrgray 0.56 & .797 & 1.02 \\  
        \end{tabular} \\ 
    \hline

  Jetson  4 GB      
        & \begin{tabular}[c]{@{}c@{}}Layer\\
                                     EC\\ 
                                     MC\\ 
                                IT\\

        \end{tabular}

        & \begin{tabular}{p{1.1cm}|p{1.1cm}|p{1.1cm}|p{1.1cm}}
         \clrgray2&4&6&8\\\hline
          \clrgray6.27& N/A & N/A & N/A\\
          \clrgray355.5&  & & \\
         \clrgray0.27&  & &  \\
        \end{tabular}

        & \begin{tabular}{p{1.1cm}|p{1.1cm}|p{1.1cm}|p{1.1cm}}
         \clrgray2&4&6&8\\\hline
        
          \clrgray6.25& N/A & N/A & N/A\\
          \clrgray359.7 &  & &\\
          \clrgray0.27 &  & &  \\
        \end{tabular}
    
        & \begin{tabular}{p{1.1cm}|p{1.1cm}|p{1.1cm}|p{1.1cm}}
         2&\clrgray4&6&8\\\hline
          N/A & \clrgray6.21 & N/A & 6.43 \\
         & \clrgray410.1 & & 532.9 \\
           & \clrgray0.45 & & 0.86 \\
        \end{tabular}
                               
        & \begin{tabular}{p{1.1cm}|p{1.1cm}|p{1.1cm}|p{1.1cm}}
                                    \clrgray 2 & 4 & 6 & 8\\\hline
                                    \clrgray6.27  & 6.31 & 6.36 & N/A\\ 
                                  \clrgray366 & 408.8 & 459.4 &\\ 
                                 \clrgray0.30 & 0.46 & 0.66 &\\ 
        \end{tabular}                         
        
        & \begin{tabular}{p{1.1cm}|p{1.1cm}|p{1.1cm}|p{1.1cm}}
                                   2 & \clrgray4 & 6 & 8\\\hline
                                N/A & \clrgray6.39 & 6.4 & 6.47\\ 
                                &\clrgray 423 & 488.4 & 524\\ 
                                & \clrgray0.50 & 0.76 & 0.87\\
        \end{tabular}  \\

    \hline
  UP$^2$ \par 2 GB   
            & \begin{tabular}{@{}c@{}}Layer\\
                                     EC\\ 
                                     MC\\ 
                                IT\\ 
            \end{tabular}
        & \begin{tabular}{p{1.1cm}|p{1.1cm}|p{1.1cm}|p{1.1cm}}
         \clrgray2&4&6&8\\\hline
          \clrgray10.897& N/A & N/A &N/A \\
          \clrgray670.9&  &  & \\
         \clrgray0.12 &  &  & \\
        \end{tabular}
                            
        & \begin{tabular}{p{1.1cm}|p{1.1cm}|p{1.1cm}|p{1.1cm}}
         \clrgray2&4&6&8\\\hline
         \clrgray10.9 & N/A & N/A&N/A \\
          \clrgray708.6&  &  & \\
          \clrgray0.12&  &  & \\
        \end{tabular} 

        & \begin{tabular}{p{1.1cm}|p{1.1cm}|p{1.1cm}|p{1.1cm}}
         2&\clrgray4&6&8\\\hline
          N/A&\clrgray 10.94 & N/A& 10.79  \\
          &\clrgray 606.8 &  & 741.7 \\
          &\clrgray 0.21 &  & 0.39 \\
        \end{tabular} 
                            
        & \begin{tabular}{p{1.1cm}|p{1.1cm}|p{1.1cm}|p{1.1cm}}
         \clrgray2&4&6&8\\\hline
          \clrgray10.96& 11.05 & 10.78&N/A \\
          \clrgray696.8& 693 & 742.6 & \\
          \clrgray0.13& 0.21 & 0.30 & \\
        \end{tabular}

        & \begin{tabular}{p{1.1cm}|p{1.1cm}|p{1.1cm}|p{1.1cm}}
                                     2&\clrgray4&6&8\\\hline
          N/A  &\clrgray 10.81 & 10.73& 10.76\\
          &\clrgray 741.1 & 707.4 & 782.1\\
          &\clrgray 0.23 & 0.31 & 0.42\\
        \end{tabular} \\ 
    \hline

  UP$^2$ \par 4 GB      
        & \begin{tabular}[c]{@{}c@{}}Layer\\
                                     EC\\ 
                                     MC\\ 
                                IT\\

        \end{tabular}

        & \begin{tabular}{p{1.1cm}|p{1.1cm}|p{1.1cm}|p{1.1cm}}
         \clrgray2&4&6&8\\\hline
          \clrgray10.17& N/A & N/A&N/A \\
         \clrgray738.3 &  &  & \\
          \clrgray0.12&  &  & \\
        \end{tabular}

        & \begin{tabular}{p{1.1cm}|p{1.1cm}|p{1.1cm}|p{1.1cm}}
        \clrgray 2&4&6&8\\\hline
        \clrgray  11.05& N/A & N/A& N/A\\
        \clrgray  653.2&  &  & \\
         \clrgray 0.12&  &  & \\
        \end{tabular}
    
        & \begin{tabular}{p{1.1cm}|p{1.1cm}|p{1.1cm}|p{1.1cm}}
         2&\clrgray4&6&8\\\hline
          N/A& \clrgray10.04 &N/A &11.51 \\
          &\clrgray 715.7 & &870 \\
          &\clrgray 0.21 & & 0.40\\
        \end{tabular}
                               
        & \begin{tabular}{p{1.1cm}|p{1.1cm}|p{1.1cm}|p{1.1cm}}
                                    \clrgray2 & 4 & 6 & 8\\\hline
         \clrgray 10.99& 11.20 & 11.38& N/A\\
          \clrgray736.8& 753.4 & 792.9 & \\
          \clrgray0.13& 0.22 & 0.30 & \\
        \end{tabular}                         
        
        & \begin{tabular}{p{1.1cm}|p{1.1cm}|p{1.1cm}|p{1.1cm}}
                                   2 &\clrgray 4 & 6 & 8\\\hline
         N/A & \clrgray11.01 & 11.38& 11.16 \\
          &\clrgray 736.2 & 798.3 & 875.3\\
          &\clrgray 0.24 & 0.30 & 0.41\\
        \end{tabular}  \\

    \hline

  UDOO \par 2 GB   
            & \begin{tabular}{@{}c@{}}Layer\\
                                     EC\\ 
                                     MC\\ 
                                IT\\ 
            \end{tabular}
        & \begin{tabular}{p{1.1cm}|p{1.1cm}|p{1.1cm}|p{1.1cm}}
         \clrgray2&4&6&8\\\hline
         \clrgray23.08 & N/A  & N/A & N/A\\
          \clrgray379.3&  &  & \\
          \clrgray0.06&  &  & \\
        \end{tabular}
                               
        & \begin{tabular}{p{1.1cm}|p{1.1cm}|p{1.1cm}|p{1.1cm}}
         \clrgray2&4&6&8\\\hline
         \clrgray 23.12&N/A   &N/A  &N/A \\
         \clrgray 424.6&  &  & \\
         \clrgray 0.06&  &  & \\
        \end{tabular}

        & \begin{tabular}{p{1.1cm}|p{1.1cm}|p{1.1cm}|p{1.1cm}}
         2&\clrgray4&6&8\\\hline
         N/A &\clrgray 23.43  & N/A &23.64 \\
          &\clrgray 361.5 &  & 422\\
          & \clrgray0.10 &  & 0.12\\
        \end{tabular} 
                            
        & \begin{tabular}{p{1.1cm}|p{1.1cm}|p{1.1cm}|p{1.1cm}}
         \clrgray2&4&6&8\\\hline
          \clrgray23.28& 23.38 & 23.62 & N/A\\
          \clrgray377.9& 348.1 & 461 & \\
          \clrgray0.06& 0.09 & 0.13 & \\
        \end{tabular}

        & \begin{tabular}{p{1.1cm}|p{1.1cm}|p{1.1cm}|p{1.1cm}}
                                     2&\clrgray4&6&8\\\hline
         N/A &\clrgray 23.57 & 23.56 & 23.46\\
          &\clrgray 360.9 & 468.2 & 453.1\\
          &\clrgray 0.10 & 0.14 & 0.17\\
        \end{tabular} \\ 
    \hline

  UDOO \par 4 GB      
        & \begin{tabular}[c]{@{}c@{}}Layer\\
                                     EC\\ 
                                     MC\\ 
                                IT\\

        \end{tabular}

        & \begin{tabular}{p{1.1cm}|p{1.1cm}|p{1.1cm}|p{1.1cm}}
         \clrgray2&4&6&8\\\hline
         \clrgray23.076 & N/A & N/A & N/A\\
         \clrgray450.1 &  &  & \\
         \clrgray0.07 &  &  & \\
        \end{tabular}

        & \begin{tabular}{p{1.1cm}|p{1.1cm}|p{1.1cm}|p{1.1cm}}
         \clrgray2&4&6&8\\\hline
        
         \clrgray23.12 & N/A & N/A & N/A\\
         \clrgray445.4 &  &  & \\
         \clrgray 0.07&  &  & \\
        \end{tabular}
    
        & \begin{tabular}{p{1.1cm}|p{1.1cm}|p{1.1cm}|p{1.1cm}}
         2&\clrgray4&6&8\\\hline
         N/A &\clrgray 22.78 &N/A  & 23.18\\
          &\clrgray 487.3 &  &596.8 \\
          &\clrgray 0.11 &  & 0.16\\
        \end{tabular}
                               
        & \begin{tabular}{p{1.1cm}|p{1.1cm}|p{1.1cm}|p{1.1cm}}
                                    \clrgray2 & 4 & 6 & 8\\\hline
          \clrgray22.98& 23.05 & 22.99 & N/A\\
          \clrgray443.8& 505.9 & 547.8 & \\
          \clrgray0.08& 0.11 & 0.15 & \\
        \end{tabular}                         
        
        & \begin{tabular}{p{1.1cm}|p{1.1cm}|p{1.1cm}|p{1.1cm}}
                                   2 &\clrgray 4 & 6 & 8\\\hline
          N/A & \clrgray23.13 & 22.96 & 23.15\\
          & \clrgray505.7 & 544.7 & 601.9 \\
          & \clrgray0.12 & 0.16 & 0.16 \\
        \end{tabular}  \\

    \hline
                                                 
    \hline

    \end{tabular}
 \end{adjustbox}
 \vspace{1mm}\caption{\label{Performancetable_sentiment1} 
 Performance of SC task for the GoEmotion dataset on various configurations.~In metrics column, EC=Energy Consumption (W), MC= Memory  (MB), and IT= Inference Time (s). The gray cells highlight the best-case scenario for each $F_1$ threshold ($\theta$).}
 \label{intentparamstable}
\end{table*}

\begin{table*}[!htb]\huge
    \centering
   \begin{adjustbox}{width=\linewidth,center}%
    \begin{tabular}{p{1.5cm}|c|c|c|c|c|c}
    \hline 
    \multirow{2}{*}{\textbf{Plat.}} & \multirow{2}{*}{\textbf{Metrics}}
    &  \multicolumn{5}{c}{\bf $\boldsymbol F_1$ Score Threshold ($\boldsymbol \theta$)} \\ \cline{3-7} & 
    & { $\boldsymbol\theta_{50}$} 
    & {$\boldsymbol\theta_{60}$} 
    & {$\boldsymbol\theta_{70}$} 
    & {{$\theta_{80}$}} 
    & {\textbf{$\theta_{90}$}} \\\hline\hline

    \hline
  Pi \par 2 GB   
            & \begin{tabular}{@{}c@{}}Layer\\
                                     EC\\ 
                                     MC\\ 
                                IT\\ 
            \end{tabular}
        & \begin{tabular}{p{1.2cm}|p{1.1cm}|p{1.0cm}|p{1.1cm}}
         \clrgray2&4&6&8\\\hline
          \clrgray4.28&N/A  &N/A &N/A \\
          \clrgray676.3&  &  & \\
          \clrgray0.55&  &  & \\
        \end{tabular}

        &\begin{tabular}{p{1.1cm}|p{1.1cm}|p{1.1cm}|p{1.1cm}}
         \clrgray2&4&6&8\\\hline
          \clrgray4.36& N/A & N/A & N/A\\
          \clrgray709& & & \\
         \clrgray0.55 &  & & \\
        \end{tabular}

        & \begin{tabular}{p{1.0cm}|p{1.2cm}|p{1.1cm}|p{1.1cm}}
         2&\clrgray4&6&8\\\hline
          N/A &\clrgray 4.3  & 4.51  & N/A \\
         &\clrgray 721.3 & 678.4 &  \\
         &\clrgray 1.08 & 0.61 &  \\
        \end{tabular} 
                            
        & \begin{tabular}{p{1.1cm}|p{1.1cm}|p{1.1cm}|p{1.1cm}}
         2&4&\clrgray6&8\\\hline
        N/A & 3.95  &\clrgray 4.35 & 4.51 \\
          & 736.8 &\clrgray 700.3 & 678.4 \\
          & 1.19 &\clrgray 0.61 & 0.61 \\
        \end{tabular}

        & \begin{tabular}{p{1.1cm}|p{1.1cm}|p{1.1cm}|p{1.1cm}}
                                     2&4&\clrgray6&8\\\hline
                                    N/A & N/A & \clrgray4.38 & 4.53 \\ 
                              &  & \clrgray704.7 & 692.2 \\ 
                              &  & \clrgray0.56 & 0.65 \\  
        \end{tabular} \\ 
    \hline

  Pi \par 4 GB      
        & \begin{tabular}[c]{@{}c@{}}Layer\\
                                     EC\\ 
                                     MC\\ 
                                IT\\

        \end{tabular}

        & \begin{tabular}{p{1.2cm}|p{1.1cm}|p{1.0cm}|p{1.1cm}}
         \clrgray2&4&6&8\\\hline
        \clrgray 4.48 & N/A & N/A & N/A\\
         \clrgray675 &  & & \\
         \clrgray0.49 &  & &  \\
        \end{tabular}

        & \begin{tabular}{p{1.1cm}|p{1.1cm}|p{1.1cm}|p{1.1cm}}
         \clrgray2&4&6&8\\\hline
        
          \clrgray4.53 &N/A  & N/A & N/A\\
          \clrgray698.3 &  & &\\
          \clrgray0.516 &  & &  \\
        \end{tabular}
    
        & \begin{tabular}{p{1.0cm}|p{1.2cm}|p{1.1cm}|p{1.1cm}}
         2&\clrgray4&6&8\\\hline
          N/A&\clrgray 4.62 &4.63 &N/A \\
         &\clrgray  683.6& 698.9& \\
           &\clrgray 0.57 & 0.57 &  \\
        \end{tabular}
                               
        & \begin{tabular}{p{1.1cm}|p{1.1cm}|p{1.1cm}|p{1.1cm}}
                                    2 &\clrgray 4 & 6 & 8\\\hline
                                    N/A  &\clrgray 4.59  & 4.85  & 4.82\\ 
                                  &\clrgray 699.1 & 706 & 686\\ 
                                 &\clrgray 0.55 & 0.57 & 0.63\\ 
        \end{tabular}                         
        
        & \begin{tabular}{p{1.1cm}|p{1.1cm}|p{1.1cm}|p{1.1cm}}
                                   2 & 4 &\clrgray 6 & 8\\\hline
                                N/A &  N/A &\clrgray 4.9 & 4.88 \\ 
                                &  & \clrgray706.6 & 709\\ 
                                &  &\clrgray 0.60 & 0.65\\
        \end{tabular}  \\

    \hline

\hline
  Jetson \par 2 GB   
            & \begin{tabular}{@{}c@{}}Layer\\
                                     EC\\ 
                                     MC\\ 
                                IT\\ 
            \end{tabular}
        & \begin{tabular}{p{1.2cm}|p{1.1cm}|p{1.0cm}|p{1.1cm}}
         \clrgray2&4&6&8\\\hline
          \clrgray5.74& N/A & N/A & N/A \\
         \clrgray314.6 &  &  & \\
          \clrgray0.29 &  &  & \\
        \end{tabular}

        &\begin{tabular}{p{1.1cm}|p{1.1cm}|p{1.1cm}|p{1.1cm}}
         \clrgray2&4&6&8\\\hline
         \clrgray5.8 & N/A & N/A & N/A \\
         \clrgray348.2 & & & \\
          \clrgray0.29 &  & & \\
        \end{tabular}

        & \begin{tabular}{p{1.0cm}|p{1.2cm}|p{1.1cm}|p{1.1cm}}
         2&4&\clrgray6&8\\\hline
          N/A & 5.96  &\clrgray 5.7  &N/A  \\
         & 388.7 &\clrgray 357.7 &  \\
         & 0.45 &\clrgray 0.29 &  \\
        \end{tabular} 
                            
        & \begin{tabular}{p{1.1cm}|p{1.1cm}|p{1.1cm}|p{1.1cm}}
         2&4&6&\clrgray8\\\hline
         N/A & 6.05 & 5.77 &\clrgray 5.7\\
          & 392.4 & 359 &\clrgray 362.7\\
          & 0.50 & 0.31 &\clrgray 0.29 \\
        \end{tabular}

        & \begin{tabular}{p{1.1cm}|p{1.1cm}|p{1.1cm}|p{1.1cm}}
                                     2&4&\clrgray6&8\\\hline
                                    N/A & N/A &\clrgray 5.8 & 5.71 \\ 
                              &  & \clrgray 362.3 & 368.9 \\ 
                              &  &\clrgray 0.33 & 0.29 \\  
        \end{tabular} \\ 
    \hline

  Jetson  4 GB      
        & \begin{tabular}[c]{@{}c@{}}Layer\\
                                     EC\\ 
                                     MC\\ 
                                IT\\

        \end{tabular}

        & \begin{tabular}{p{1.2cm}|p{1.1cm}|p{1.0cm}|p{1.1cm}}
         \clrgray2&4&6&8\\\hline
         \clrgray5.74 & N/A  & N/A & N/A \\
         \clrgray368.3  &  & & \\
         \clrgray0.29&  & &  \\
        \end{tabular}

        & \begin{tabular}{p{1.1cm}|p{1.1cm}|p{1.1cm}|p{1.1cm}}
         \clrgray2&4&6&8\\\hline
        
         \clrgray5.76 & N/A  & N/A & N/A \\
         \clrgray365.8 &  & &\\
          \clrgray0.27 &  & &  \\
        \end{tabular}
    
        & \begin{tabular}{p{1.0cm}|p{1.2cm}|p{1.1cm}|p{1.1cm}}
         2&4&\clrgray6&8\\\hline
         N/A & 6.23 &\clrgray 5.95 & N/A \\
         & 418.5 &\clrgray 368.7 & \\
           & 0.497 & \clrgray0.29 &  \\
        \end{tabular}
                               
        & \begin{tabular}{p{1.1cm}|p{1.1cm}|p{1.1cm}|p{1.1cm}}
                                    2 & 4 & 6 &\clrgray 8\\\hline
                                     N/A & 6.04  & 6.12 &\clrgray 6.04\\ 
                                  & 424.4 & 365.5 &\clrgray 361.7\\ 
                                 & 0.45 & 0.33 &\clrgray 0.29\\ 
        \end{tabular}                         
        
        & \begin{tabular}{p{1.1cm}|p{1.1cm}|p{1.1cm}|p{1.1cm}}
                                   2 & 4 &\clrgray 6 & 8\\\hline
                                 N/A & N/A &\clrgray 6.05 & 6.08\\ 
                                &  &\clrgray 363.5 & 366.6\\ 
                                &  &\clrgray 0.32 & 0.30\\
        \end{tabular}  \\

    \hline

  UP$^2$ \par 2 GB   
            & \begin{tabular}{@{}c@{}}Layer\\
                                     EC\\ 
                                     MC\\ 
                                IT\\ 
            \end{tabular}
        & \begin{tabular}{p{1.2cm}|p{1.1cm}|p{1.0cm}|p{1.1cm}}
         \clrgray2&4&6&8\\\hline
          \clrgray11.108& N/A & N/A & N/A \\
          \clrgray512.5&  &  & \\
          \clrgray0.13&  &  & \\
        \end{tabular}

        &\begin{tabular}{p{1.1cm}|p{1.1cm}|p{1.1cm}|p{1.1cm}}
         \clrgray2&4&6&8\\\hline
          \clrgray11.2& N/A & N/A & N/A \\
          \clrgray597.6&  &  & \\
          \clrgray0.12&  &  & \\
        \end{tabular}

        & \begin{tabular}{p{1.0cm}|p{1.2cm}|p{1.1cm}|p{1.1cm}}
         2&4&\clrgray6&8\\\hline
          N/A&11.18 &\clrgray 11.44 & N/A \\
          & 652.9 &\clrgray 574 & \\
          & 0.19 &\clrgray 0.18 & \\
        \end{tabular} 
                            
        & \begin{tabular}{p{1.1cm}|p{1.1cm}|p{1.1cm}|p{1.1cm}}
         2&4&\clrgray6&8\\\hline
          N/A& 11.21&\clrgray 10.77 & 10.72 \\
          & 650.8 &\clrgray 601.8 & 603.5\\
          & 0.20 &\clrgray 0.12 & 0.12\\
        \end{tabular}

        & \begin{tabular}{p{1.1cm}|p{1.1cm}|p{1.1cm}|p{1.1cm}}
                                     2&4&\clrgray6&8\\\hline
          N/A& N/A&\clrgray 10.63 &  10.7 \\
          &  &\clrgray 591.3 & 591.1\\
          &  &\clrgray 0.12 & 0.13\\
        \end{tabular} \\ 
    \hline

  UP$^2$ \par  4 GB      
        & \begin{tabular}[c]{@{}c@{}}Layer\\
                                     EC\\ 
                                     MC\\ 
                                IT\\

        \end{tabular}

        & \begin{tabular}{p{1.2cm}|p{1.1cm}|p{1.0cm}|p{1.1cm}}
         \clrgray2&4&6&8\\\hline
          \clrgray10.99&N/A & N/A & N/A \\
          \clrgray599.8&  &  & \\
          \clrgray0.12&  &  & \\
        \end{tabular}

        & \begin{tabular}{p{1.1cm}|p{1.1cm}|p{1.1cm}|p{1.1cm}}
         \clrgray2&4&6&8\\\hline
        
          \clrgray11.21&N/A & N/A & N/A \\
          \clrgray601.8&  &  & \\
          \clrgray0.11&  &  & \\
        \end{tabular}
    
        & \begin{tabular}{p{1.0cm}|p{1.2cm}|p{1.1cm}|p{1.1cm}}
         2&\clrgray4&6&8\\\hline
          N/A&\clrgray 11.36& 11.33 & N/A \\
          &\clrgray  651.4& 658.1 & \\
          &\clrgray  0.19& 0.20 & \\
        \end{tabular}
                               
        & \begin{tabular}{p{1.1cm}|p{1.1cm}|p{1.1cm}|p{1.1cm}}
                                    2 &\clrgray 4 & 6 & 8\\\hline
          N/A&\clrgray 11.32& 11.40 & 11.35 \\
          &\clrgray 653.7 & 652.7 & 657.7\\
          &\clrgray 0.19 & 0.19 & 0.22 \\
        \end{tabular}                         
        
        & \begin{tabular}{p{1.1cm}|p{1.1cm}|p{1.1cm}|p{1.1cm}}
                                   2 & 4 &\clrgray 6 & 8\\\hline
          N/A& N/A&\clrgray 11.16 & 11.3 \\
          &  &\clrgray 657.7 & 652.9\\
          &  &\clrgray 0.22 & 0.21\\
        \end{tabular}  \\

    \hline

  UDOO \par 2 GB   
            & \begin{tabular}{@{}c@{}}Layer\\
                                     EC\\ 
                                     MC\\ 
                                IT\\ 
            \end{tabular}
        & \begin{tabular}{p{1.2cm}|p{1.1cm}|p{1.0cm}|p{1.1cm}}
         \clrgray2&4&6&8\\\hline
          \clrgray24.02& N/A  & N/A  & N/A  \\
          \clrgray437.8&  &  & \\
          \clrgray0.07&  &  & \\
        \end{tabular}

        &\begin{tabular}{p{1.1cm}|p{1.1cm}|p{1.1cm}|p{1.1cm}}
         \clrgray2&4&6&8\\\hline
         \clrgray24.08 & N/A  & N/A  &  N/A \\
          \clrgray438.6&  &  & \\
          \clrgray0.05&  &  & \\
        \end{tabular}

        & \begin{tabular}{p{1.0cm}|p{1.2cm}|p{1.1cm}|p{1.1cm}}
         2&4&\clrgray6&8\\\hline
           N/A& 24.09 &\clrgray24.07  &  N/A \\
          &491.2 &\clrgray 490.8 & \\
          &0.09 &\clrgray 0.09 & \\
        \end{tabular} 
                            
        & \begin{tabular}{p{1.1cm}|p{1.1cm}|p{1.1cm}|p{1.1cm}}
         2&4&\clrgray6&8\\\hline
           N/A&24.14 &\clrgray 24.06 & 24.02 \\
          &  490.9&\clrgray 490.5 & 503.4\\
          & 0.09 &\clrgray 0.09 & 0.09\\
        \end{tabular}

        & \begin{tabular}{p{1.1cm}|p{1.1cm}|p{1.1cm}|p{1.1cm}}
                                     2&4&6&\clrgray8\\\hline
           N/A&  N/A &24.11 &\clrgray 24.07 \\
          &  &490.3 &\clrgray 492.4\\
          &  &0.09 &\clrgray 0.08\\
        \end{tabular} \\ 
    \hline

  UDOO \par  4 GB      
        & \begin{tabular}[c]{@{}c@{}}Layer\\
                                     EC\\ 
                                     MC\\ 
                                IT\\

        \end{tabular}

        & \begin{tabular}{p{1.2cm}|p{1.1cm}|p{1.0cm}|p{1.1cm}}
         \clrgray2&4&6&8\\\hline
          \clrgray22.79&  &  &  \\
          \clrgray445.6&  &  & \\
          \clrgray0.06&  &  & \\
        \end{tabular}

        & \begin{tabular}{p{1.1cm}|p{1.1cm}|p{1.1cm}|p{1.1cm}}
         \clrgray2&4&6&8\\\hline
        
          \clrgray22.86&  &  &  \\
          \clrgray443&  &  & \\
          \clrgray0.05&  &  & \\
        \end{tabular}
    
        & \begin{tabular}{p{1.0cm}|p{1.2cm}|p{1.1cm}|p{1.1cm}}
         2&\clrgray4&6&8\\\hline
          &\clrgray 23.07 &23.57  &  \\
          &\clrgray 500.1 & 496.4 & \\
          &\clrgray 0.08 & 0.09 & \\
        \end{tabular}
                               
        & \begin{tabular}{p{1.1cm}|p{1.1cm}|p{1.1cm}|p{1.1cm}}
                                    2 &\clrgray 4 & 6 & 8\\\hline
          &\clrgray 23.09 & 23.49 & 23.39 \\
          &\clrgray 497.5 & 499.5 & 510.1\\
          &\clrgray 0.09 & 0.08 &0.09 \\
        \end{tabular}                         
        
        & \begin{tabular}{p{1.1cm}|p{1.1cm}|p{1.1cm}|p{1.1cm}}
                                   2 & 4 &\clrgray 6 & 8\\\hline
          &  &\clrgray 23.58 & 23.50 \\
          &  &\clrgray 495.2 & 499.8\\
          &  &\clrgray 0.08 & 0.08\\
        \end{tabular}  \\

    \hline                                              
    \hline

    \end{tabular}
 \end{adjustbox}
 \vspace{1mm}
 \caption{\label{Performancetable_ner} 
 Performance of NER task for the WNUT17 dataset on various configurations.~In metrics column, MS= Model Size (MB), Params= Parameters (Million), EC=Energy Consumption (Watt), MC= Memory  (MB), and IT= Inference Time (s). The gray cells highlight the best-case scenario for each $F_1$ threshold ($\theta$).}
 \label{Performancetable_NER1}
\end{table*}

\subsubsection{Observations}\label{sec:observations}

We now discuss our major observations and address the research questions introduced in Sec.~\ref{sec:rq}.


\paragraph{\textbf{Selecting ``suitable'' model subject to given constraints [RQ 1].}}

We can address this specific research question by inspecting Table~\ref{Performancetable_sentiment1} and ~\ref{Performancetable_NER1}. Note that Table~\ref{Performancetable_sentiment1} and ~\ref{Performancetable_NER1} provide information on the model size, performance, parameters, and pruning for the SC and NER tasks, respectively. Let us assume a system designer is looking for suitable NER models for a 2 GB embedded platform that maintains approximately 70\% of BERT's accuracy ($\theta_{70}$). In this case, we can \ca scan through the NER performance metrics (\ie Table \ref{Performancetable_NER1}), and \cb observe from Pi 2~GB \textit{Platform} row and $\theta_{70}$ column that a six-layered and pruned (45\% reduced) BERT model can run on a Pi 2~GB platform and attain 70\% of $BERT_{Base}$'s original $F_1$ score. Hence, our exploration (and similar experiments along this line) can \textit{aid the designers to select appropriate models with desired performance guarantees}.


 
 

\paragraph{\textbf{Accuracy and model-size trade-offs for pruned architectures [RQ 2].}}

Table~\ref{Performancetable_sentiment1} and Table~\ref{Performancetable_NER1} further provide insights on the pruning vs. $F_1$ score trade-off. For example, in Table~\ref{Performancetable_sentiment1}, the Pi 2~GB \textit{Platform} row shows a set of models that can run on that system. The same row and $\theta_{80}$ (80\% \textit{$F_1$ score threshold}) column show  that even pruning 55\% of a $BERT_{Base}$ model with four layers can retain 80\% of original $F_1$ score of $BERT_{Base}$, while the model size can be reduced to 198.7~MB from 441.55~MB. Tables \ref{Performancetable_sentiment1} and \ref{Performancetable_NER1} indicate that although \textit{pruning has only a minor impact on memory consumption, it does not have a significant effect on energy consumption}. Furthermore, our analysis of Table \ref{Performancetable_sentiment1} suggests that \textit{inference time is directly proportional to the size of the model, implying that decreasing the model size leads to a decrease in inference time.}



\paragraph{\textbf{Accuracy vs. system resource trade-offs for pruned architectures [RQ 3].}}

If a user has precise requirements for inference time and memory consumption for a given hardware, one can scan Table ~\ref{Performancetable_sentiment1} and ~\ref{Performancetable_NER1} to pick the optimum model that meets those requirements. For instance, if we want to find NER models that can make inferences in less than 0.56 seconds on Raspberry Pi that has 4~GB of memory, the corresponding  \textit{Platform} row Pi (4~GB) in Table~\ref{Performancetable_NER1}, shows us the model parameters that can satisfy this requirement(\eg two-layered, four-layered). Since both of them are feasible for the chosen platform, designers can choose any of them based on the required application performance. As an example, if we pick a two-layered BERT model, the accuracy is 60\%, and the memory consumption is 698.3~MB. In contrast, if we select the four-layered BERT model, it can achieve 80\% accuracy, with a cost of higher memory consumption of 699.1 MB. Hence, at the expense of slightly higher memory consumption, it is possible to get 20\% more accuracy. Such a lookup-based approach allows the designers to perform a desired \textit{cost-benefit analysis}. 
\vspace{-0.1cm}

\paragraph{\textbf{The case for GPUs [RQ 4].}}


Intuitively, GPUs aid in any learning-enabled tasks. We used one GPU-enabled hardware (\jetn) in our design space exploration. As Table~\ref{Performancetable_sentiment1} and Table~\ref{Performancetable_NER1} illustrate \jet reduces inference time compared to the \rpi board (no GPU). However, GPUs alone cannot provide faster inference. For instance, x86 boards (\up and \udo) do not have GPUs but are equipped with a faster processor, and hence result in better inference times (\ie took less time to process the queries). Systems with better hardware (CPU/GPU/memory) can output inference decisions faster, which may increase power consumption, as we discuss next.


\begin{figure}[!t]
	\centering
	\includegraphics[width=1.03\linewidth]{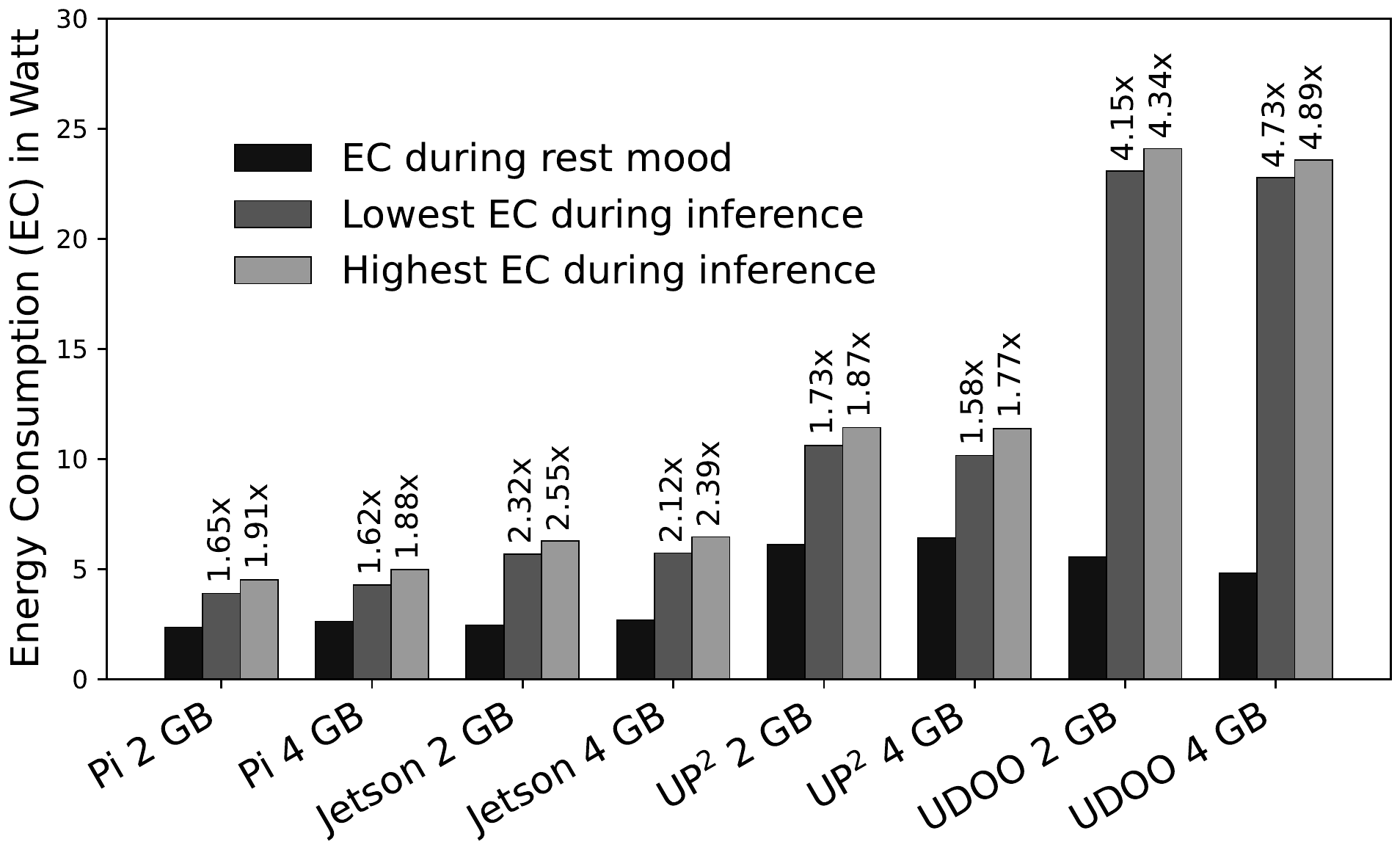}
	\caption{Energy consumption during rest mode and inference period. ARM devices use less energy compared to x86 systems.}
	\label{fig:energy_consum}
\end{figure}

\paragraph{\textbf{Energy consumption on various architectures and model configurations [RQ 5].}}

Since many embedded platforms used for NLP tasks (\eg voice-controlled robots, voice assistants, IoT devices) are battery-operated, energy consumption for inferring user commands is a crucial parameter. Hence, we also analyze the energy usage of the NLP tasks. For any selected BERT model, one can find the system energy consumption from Table~\ref{Performancetable_sentiment1} and ~\ref{Performancetable_NER1}, for two different tasks, respectively. 
As the table shows, (for a given hardware) during the inference of a given command, \textit{energy consumption does not vary significantly for various models}.

To understand the energy usage of the various NLP tasks on our test platforms, we also measured the energy consumption of each board during rest mode and inference period (see Fig.~\ref{fig:energy_consum}). 
We obtained the rest mode energy usage by idling the device for 10 minutes and took the average value. For inference energy consumption, we tested with 40 SC and 40 NER queries and repeated each of them 100 times (\ie a total of $2\times 40\times 100 = 8000$ samples. We report the maximum (lighter gray) bar and minimum (darker gray) energy consumption values of the 8000 trials. 

As Fig.~\ref{fig:energy_consum} shows, the inference energy consumption increases by a factor of 1.73 to 4.89 times compared to the rest mode energy usage. Besides, ARM architectures (\rpi and \jet) consume less energy than the x86 architectures (\up and \udo). Another interesting observation is that even though \jet boards use GPU, they are more power efficient for performing NLP tasks compared to some CPU-only x86 systems. Our experiments show that the AMD Ryzen platform (\udo) performs poorly in terms of energy usage. However, as Table~\ref{Performancetable_sentiment1} and Table~\ref{Performancetable_ner} indicate, \udo boards output faster inference time (since they have relatively faster CPU than the others). Hence, there exists a \textit{trade-off} between inference time and energy usage.

\subsubsection{Summary of Findings} 

Our key findings for custom BERT architectures are listed below.

{\centering
\begin{tcolorbox}[colback=gray!20,colframe=black, boxrule=0.0pt, width=\columnwidth, left=0.5pt,
	right=0.5pt,
	top=0.5pt,
	bottom=0.5pt]
\textbf{Model Size \& Pruning.}
\begin{itemize}[leftmargin=*]
    \item Pruning helps in reduction in size (upto 67\%) while maintaining at least 50\% of $BERT_{Base}$'s F1 score.
    \item The time required for inference is directly related to the size of the model, \ie a smaller model size results in a reduction in inference time.
    \item Pruning of attention heads does not reduce memory usage.
    \item Pruning attention heads does not improve energy consumption significantly.
\end{itemize}
\end{tcolorbox} %
}

{\centering
\begin{tcolorbox}[colback=gray!20,colframe=black, boxrule=0.0pt, width=\columnwidth, left=0.5pt,
	right=0.5pt,
	top=0.5pt,
	bottom=0.5pt]
\textbf{System Artifacts.}
\begin{itemize}[leftmargin=*]
	\item Faster x86 (Intel and AMD) platforms outperform ARM SoCs (\eg Pi and Jetson boards) wrt. inference time, but their energy consumption is significantly higher (\eg at least 2.60 times) than ARM counterparts.
	\item GPUs aid in performance (\eg inference times are approx. 2-times faster in \jet than \rpi) but GPUs alone in \jet boards cannot outperform a relatively faster CPU (\ie those used in \up and \udo boards).
	\item Powerful processor can decrease inference time (as expected) but comes with a cost (increased power consumption: 2.60-5.90 times higher).
\end{itemize}
\end{tcolorbox} %
}


\section{Discussion}\label{discussion}

 We explore different custom architectures of BERT-based language models and test their deployment feasibility in low-power embedded devices. We conducted extensive performance evaluations on four embedded platforms from various vendors with varying computing capabilities to cover a wide range of application scenarios.
 We show that it is \textit{not always feasible to shrink} the size of \enquote{finetuned} $BERT_{Base}$ model 
that can satisfy specific user-defined accuracy/performance budgets. 
We also report which models are deployable to resource-constrained devices with given user requirements. We believe our empirical findings will help the developers quickly narrow down the plausible BERT-based architecture for target applications, thus saving development time and effort. While we tested the NLP models on four embedded platforms (a total of 8 hardware configurations) in a Linux environment, they can be ported to other systems, such as smartphones/tablets running different OSes (such as Android). Thus our empirical study is applicable in broader human-centric application domains, including chatbots~\cite{adamopoulou2020chatbots, brandtzaeg2017people, shum2018eliza}, virtual assistants~\cite{white2018skill, schmidt2018industrial}, and language translation~\cite{oettinger2013automatic, brown1988statistical}.

Our study is limited to BERT-based models for four existing datasets (\ie may not generalize to other language models and datasets). However, our evaluation framework is modular and can be retrofitted to other architectures/datasets without loss of generality. While shrinking models have made it possible to deploy them on resource-constrained embedded devices, their performance on new datasets or tasks is often limited. One potential solution to mitigate this issue is to utilize continual learning techniques~\cite{de2021continual, hadsell2020embracing}, as they allow models to continuously learn and evolve based on increasing data input while retaining previously acquired knowledge. Our future work will explore the feasibility of employing continual learning for embedded devices.

One of the challenges to figuring out the \textit{optimal} BERT-based architecture is the lack of application-specific (\viz voice-controlled robots for home automation) datasets. Existing datasets either \ca do not have enough examples for training deep learning models or \cb do not provide complex, practical queries to test the robustness of a given model. Building suitable datasets for IoT-specific human-centric applications such as voice-control home/industrial automation is an interesting open research problem.

\section{Related Work} \label{related_work}

 We discuss related research on two fronts: 
\ca  BERT-based models and their efficient variants and \cb using NLP on embedded devices.


\subsection{BERT-based Models and their Variants}\label{bert_study}
The performance of BERT comes at a high computation and memory cost, which makes on-device inference really challenging. 
To mitigate this issue, researchers have proposed knowledge distillation approaches from the original BERT model, for example, \ca ``finetune'' the BERT model to improve task-specific knowledge distillation~\cite{DBLP:journals/corr/abs-1908-08962, DBLP:conf/emnlp/TsaiRJALA19}, \cb use Bi-LSTM models~\cite{DBLP:journals/corr/abs-1903-12136} for knowledge distillation from BERT, \cc leverage single-task models to teach a multi-task model~\cite{DBLP:conf/acl/ClarkLKML19}, \cd distillation of knowledge from an ensemble of BERT into a single BERT~\cite{DBLP:conf/acl/LiuWJCZAHCPCG20}, \ce TinyBERT~\cite{tinybert} uses a layer-wise distillation strategy for BERT in both the pre-training and fine-tuning stages, and \cf DistilBERT~\cite{distilbert} halves the depth of the BERT model through knowledge  distillation in the pre-training stage and an optional fine-tuning stage. On a different direction, the \textit{Patient Knowledge Distillation} approach~\cite{DBLP:conf/emnlp/SunCGL19} compresses an original large model (``teacher'') into an equally-effective lightweight shallow network (``student''). Other BERT models (\eg SqueezeBERT~\cite{SqueezeBERT}, MobileBERT~\cite{mobilebert}, Q8BERT~\cite{DBLP:conf/nips/ZafrirBIW19}, ALBERT~\cite{ALBERT}) can also reduce resource consumption than the vanilla BERT. EdgeBERT~\cite{EdgeBERT}, an algorithm-hardware co-design approach, performs latency-aware energy optimizations for multi-task NLP problems. However, unlike ours, EdgeBERT \ca does not apply attention heads pruning, and \cb does not report scores on downstream NLP tasks on real-world embedded systems. 

\subsection{NLP for Embedded Platforms} 


Researchers have explored NLP techniques to facilitate natural communication between humans and embedded devices, especially in the context of voice-controlled cognitive robots. For example, \citet{rajeshkannvoicerecog} presents a voice recognition tool that compares the user's input commands with the stored data.~\citet{zhangdualarm} propose a ROS-based robot that analyzes commands using an offline grammar recognition library.~\citet{rajeshoptimalnlp} propose a cost-efficient speech processing module running on ROS that can provide natural language responses to the user. There also exists ROS-integrated independent speech recognition packages~\cite{zhangwheelchair,HRIvoice} as well as Arduino-based~\cite{robotarduino} and custom~\cite{robotsignal} voice-control robot platforms. \citet{voicebot} a voice-controlled robotic arm (named VoiceBot) for individuals with motor impairments~\cite{vj}. However, most of these works focused on rule-based approaches, and we note that transformer architectures are still under-explored in terms of their practical deployment challenges in real-world embedded and robotic devices, which is the focus of this study. 



\subsection{Uniqueness of Our Work}

While existing work can reduce the size of BERT models through distillation and pruning, from a system design perspective, it is still difficult and tedious for a developer to find out the ``right'' BERT-based architecture to use in an embedded platform. To date, it is also unclear which lighter version of BERT would find the optimal balance between the resources available in an embedded device (\eg CPU, GPU, memory) and the minimum accuracy desired. We used four off-the-shelf platforms widely used by developers for various IoT and embedded applications and benchmarked state-of-the-art BERT architectures. Our empirical evaluation and design space exploration on heterogeneous platforms (\eg x86 and ARM, with or without GPU) can help the system and machine learning engineers to pick suitable architectures depending on target system configuration and performance constraints (\eg accuracy, $F_1$ score). To the best of our knowledge, this work is \textit{one of the first efforts to study the feasibility of deploying BERT-based models} in real-world resource-constrained embedded platforms.

\section{Conclusion}\label{conclusion}

This paper presents an empirical study of BERT-based neural architectures in terms of the feasibility of deploying them on resource-constrained systems, which have become ubiquitous nowadays. Our performance evaluation results will assist developers of multiple ubiquitous computing domains, such as voice-controlled home and industrial automation, precision agriculture, and medical robots to determine the deployability of NLP models in their target platform. By using our benchmark data, designers of ubiquitous systems will now be able to select the \enquote{right} hardware, architecture, and parameters depending on the resource constraints and performance requirements. This will also save time on the developer's end, as they can make informed choices regarding which BERT-based architecture to use during development based on their NLP application scenario and the available hardware.

\begin{acks}
This work has been partially supported by the National Science Foundation Standard Grant Award 2302974, Air Force Office of Scientific Research Grant/Cooperative Agreement Award FA9550-23-1-0426, and Washington State University Grant PG00021441. Any findings, opinions, recommendations, or conclusions expressed in the paper are those of the authors and do not necessarily reflect the views of sponsors.
\end{acks}


\bibliographystyle{IEEEtranN}
\balance
\bibliography{sample-base}


\appendix

\section*{Appendix}



 \section{Dataset}\label{appen:datasets}

The details of the datasets used in our evaluations are presented below.

\subsection{Human Robot Interaction Corpus (HuRIC)} \label{huric_appen}

To utilize the HuRIC dataset in our task, we extracted intent information from \textit{frame semantics}. The original dataset\footnote{https://github.com/crux82/huric.} contains many intents that are closely related semantically; therefore, after analyzing the commands, we merged several similar intents. For instance, based on most of the commands in the dataset, we consolidated intent \enquote{\textit{Releasing}} and intent \enquote{\textit{Placing}} into a single intent, namely \enquote{\textit{Placing}}. The HuRIC dataset consists of both simple and complex commands. To illustrate the difference, consider commands like ``Bring the book on the table in the kitchen,'' which has a single intent, \enquote{\textit{Bring}}, while commands like ``Please go to the kitchen and inspect the sink'' have two different intents: \enquote{\textit{Motion}} and \enquote{\textit{Inspect}}. However, the dataset contains very few examples of complex commands.


\subsection{GoEmotion}\label{goemotion_appen}

In contrast to the basic six emotions, which include only \emph{one} positive emotion (joy), the GoEmotion taxonomy includes \emph{twelve} positive emotions. Additionally, the taxonomy includes \emph{eleven} negative, \emph{four} ambiguous emotion categories, and \emph{one} neutral emotion, making it widely suitable for conversation understanding tasks that require a subtle differentiation between emotion expressions.

The goal was to build a large dataset focused on conversational data, where emotion is a critical communication component. Because the Reddit platform offers a large, publicly available volume of content that includes direct user-to-user conversation, it is a valuable resource for emotion analysis. So, the authors built GoEmotion using Reddit comments from 2005 (the start of Reddit) to January 2019, sourced from subreddits with at least 10K comments, excluding deleted and non-English comments. The authors created a taxonomy to maximize three objectives: \ci a greater coverage of the emotions expressed in Reddit data, \cii the coverage of types of emotional expressions, and \ciii limit the overall number of emotions and their overlap. Such a taxonomy allows data-driven, fine-grained emotion understanding while addressing potential data sparsity for some emotions.


The authors iteratively defined and refined emotion label categories during the data labeling stages, considering 56 emotion categories. They removed emotions scarcely selected by raters, those with low inter-rater agreement due to similarity to other emotions or difficulty detecting from the text. Additionally, they incorporated emotions frequently suggested by raters and were well-represented in the data. Finally, they optimized emotion category names to maximize interpretability, resulting in a high inter-rater agreement, with at least two raters concurring on at least one emotion label for 94\% of the examples.

\subsection{CoNLL}\label{conll_appen}

We extracted ten days' worth of data from the files representing the end of August 1996 for the training and development set. The texts used for the test set were from December 1996, while the pre-processed raw data covers the month of September 1996. Each data file contains one word per line; empty lines represent sentence boundaries. Additionally, a tag at the end of each line indicates whether the current word is inside a named entity. The data encompasses entities of four types: persons (PER), organizations (ORG), locations (LOC), and miscellaneous names (MISC).


\subsection{WNUT 2017}\label{wnut_appen}
As discussed in \ref{subsec:ner-dataset}, user comments were mined from different social media platforms. Below, we provide some details about the dataset.
The researchers who proposed the dataset drew documents from various English-speaking subreddits over January-March 2017. They selected these based on volume, considering a variety of regions and granularities. For instance, they included country- and city-level subreddits, along with non-geo-specific forums such as \texttt{/r/homeassistant}, \texttt{r/HomeImprovement}, and \texttt{/r/restaurants}. The documents were filtered to include only those between 20 and 400 characters. Then, they split them into sentences and tagged them with the NLTK~\cite{nltk} and Stanford CoreNLP~\cite{corenlp}.

The corpus contains comments from online video-sharing platforms (\eg YouTube), specifically drawn from the all-time top 100 videos across all categories within certain parts of the anglosphere. Researchers collected one hundred top-level comments from each video, excluding non-English comments. Twitter samples were collected from periods aligning with recent natural disasters, specifically the Rigopiano avalanche and the Palm Sunday shootings. The intention was to select content about emerging events that may contain highly specific and novel toponyms.
The dataset also includes another set of user-generated content from StackExchange. Specifically, they downloaded title posts and comments associated with five topics (movies, politics, physics, sci-fi, and security) posted between January and May 2017 from \texttt{archive.org}. Finally, they uniformly selected 400 samples from these title posts and comments for each topic.

    
    

\section{Hardware Cost}\label{estimated_cost}

Table~\ref{tab:board-costs-updated} summarizes the hardware costs.

\begin{table}[!h]
    \centering
    \begin{tabular}{p{4.0cm} r}
    \toprule
    Hardware & Cost (USD) \\
    \midrule
    Raspberry Pi 2 GB & 94 \\
    Raspberry Pi 4 GB & 109 \\
    Jetson 2 GB & 159 \\
    Jetson 4 GB & 219 \\
    UP$^2$ 2 GB & 179 \\
    UP$^2$ 4 GB & 199 \\
    UDOO (including RAM) & 610 \\
    UM25C energy meter & 28 \\
    Misc. (cables, memory cards) & 100 \\
    \midrule
    Total & \$1697 \\
    \bottomrule
    \end{tabular}
    \caption{Hardware costs for setting up our experiments.}
    \label{tab:board-costs-updated}
\end{table}
\end{document}